%% file: main.tex
\documentclass[10pt,twocolumn,letterpaper]{article}
\usepackage[accsupp]{axessibility}
\usepackage[pagenumbers]{cvpr}

\input{preamble}
\definecolor{cvprblue}{rgb}{0.21,0.49,0.74}
\usepackage[pagebackref,breaklinks,colorlinks,allcolors=cvprblue]{hyperref}

\usepackage{colortbl}
\usepackage{algorithm}
\usepackage{algpseudocode}
\usepackage{adjustbox}
\usepackage{subcaption}
\usepackage{multicol}
\usepackage{multirow}
\usepackage[normalem]{ulem}
\usepackage{soul}
\usepackage{placeins}
\usepackage{float}
\setul{0.3ex}{0.2ex} %

\definecolor{cellcol}{gray}{.94}
\newcommand{\graycell}[1]{\cellcolor{cellcol}{#1}}

\def\paperID{5873} %
\def\confName{CVPR}
\def\confYear{2025}

\title{Question-Aware Gaussian Experts for Audio-Visual Question Answering} %

\author{
Hongyeob Kim\textsuperscript{1}\thanks{Equal Contribution.} \hspace{0.25em}
Inyoung Jung\textsuperscript{1}\footnotemark[1] \hspace{0.25em}
Dayoon Suh\textsuperscript{2}  \hspace{0.25em}
Youjia Zhang\textsuperscript{1}  \hspace{0.25em}
Sangmin Lee\textsuperscript{1}  \hspace{0.25em}
Sungeun Hong\textsuperscript{1}\thanks{Corresponding Author.} \\
\textsuperscript{1}Sungkyunkwan University \quad
\textsuperscript{2}Purdue University
}

\begin{document}

\maketitle
\input{sec/0_abstract}

\input{sec/1_intro}

\input{sec/2_related_works}
\input{sec/3_methods}

\input{sec/4_experiments}
\input{sec/5_ablation}

\input{sec/6_conclusion}

\newpage
{
\small
\bibliographystyle{ieeenat_fullname}
\bibliography{main}
}

\clearpage
\input{sec/X_suppl}

\end{document}

%% file: sec/0_abstract.tex
\begin{abstract}

Audio-Visual Question Answering (AVQA) requires not only question-based multimodal reasoning but also precise temporal grounding to capture subtle dynamics for accurate prediction. However, existing methods mainly use question information implicitly, limiting focus on question-specific details. Furthermore, most studies rely on uniform frame sampling, which can miss key question-relevant frames. Although recent Top-K frame selection methods aim to address this, their discrete nature still overlooks fine-grained temporal details. This paper proposes \textbf{QA-TIGER}, a novel framework that explicitly incorporates question information and models continuous temporal dynamics. Our key idea is to use Gaussian-based modeling to adaptively focus on both consecutive and non-consecutive frames based on the question, while explicitly injecting question information and applying progressive refinement. We leverage a Mixture of Experts (MoE) to flexibly implement multiple Gaussian models, activating temporal experts specifically tailored to the question. Extensive experiments on multiple AVQA benchmarks show that QA-TIGER consistently achieves state-of-the-art performance. Code is available at \hypersetup{urlcolor=magenta}\url{https://aim-skku.github.io/QA-TIGER/}

\end{abstract}

%% file: sec/1_intro.tex
\section{Introduction}
\label{sec:intro}

\input{figure/teaser}

Audio-Visual Question Answering (AVQA) focuses on analyzing and interpreting both audio and visual cues to provide accurate answers to questions.
Recent advancements in Audio-Visual Question Answering (AVQA) have focused on spatial and temporal grounding~\cite{clip-tass24,music-avqa22,valor23}, parameter-efficient models using vision transformers~\cite{dg-sct24,lavish23,stg-cma23}, and bias adjustments for improved generalization~\cite{music-avqa-r25,coca23}.
Despite promising results, many existing methods struggle to capture fine-grained, question-specific details and temporal cues essential for effective reasoning. We argue that two key considerations are critical for successful AVQA: \textit{(i) flexibly capturing and integrating question-relevant audio-visual cues across temporal contexts}, and \textit{(ii) embedding question context explicitly within the audio-visual feature processing stages}.

Most AVQA methods use uniform sampling or discrete frame selection~\cite{music-avqa22, lavish23, stg-cma23}, often overlooking question-specific details. These methods treat frames equally missing important, question-relevant information in both audio and visual modalities. Recent approaches, such as PSTP~\cite{pstp23} and TSPM~\cite{tspm24}, use Top-K frame selection to improve context relevance by aligning frames with the question. {However, they select them based solely on visual cues, relying on discrete sampling that disrupts continuity by ignoring temporal cues. Additionally, this approach focuses mainly on visual alignment, missing crucial audio details.}

In terms of question integration, most AVQA models incorporate question information only at the final reasoning stage, typically by simple multiplication~\cite{music-avqa22,lavish23,AVMamba24,tspm24}. This late-stage integration limits the model's ability to encode question-specific features during intermediate steps, reducing the effectiveness of reasoning. Although several methods attempt to enhance reasoning by using question details in frame selection~\cite{pstp23, tspm24}, this is done indirectly, and they fail to explicitly embed question context within the audio-visual encoding pipeline. This limited approach restricts the model's ability to focus on relevant temporal cues progressively.

In this paper, we propose QA-TIGER (Question-Aware Temporal Integration of Gaussian Experts for Reasoning), as illustrated in Figure~\ref{teaser}. To address issue \textit{(i)}, QA-TIGER introduces a multi-Gaussian weighting mechanism embedded within a Mixture of Experts (MoE) framework~\cite{moe17}. This approach assigns adaptive weights across both consecutive and non-consecutive temporal spans, thereby enabling the model to capture complex temporal dependencies more effectively. Gaussian experts are adaptively activated with well-positioned centers, reducing redundancy and simultaneously improving temporal alignment. In contrast to previous methods~\cite{pstp23, tspm24}, which strictly align audio cues with corresponding visual frames, QA-TIGER provides explicit temporal grounding for both audio and visual modalities independently. This allows for more accurate and robust alignment with the question-relevant segments.

To address issue \textit{(ii)}, QA-TIGER incorporates question information early in the process, creating question-aware features that permeate both the visual and audio modalities. By embedding question context directly into each modality’s features, QA-TIGER aligns with the question context throughout the entire pipeline, unlike prior methods that add question information only at the final stage~\cite{music-avqa22, lavish23, pstp23, tspm24}. This approach allows the model to progressively refine its focus based on the question, resulting in deeper and more explicit integration of question relevance. Ultimately, this approach enhances accuracy and contextual alignment across all reasoning stages.

Our contributions can be summarized as follows:
\begin{itemize} 
\item We propose QA-TIGER, a framework that adaptively models temporal dynamics using strategically positioned Gaussian experts, capturing question-relevant information across continuous temporal spans and adjusting segment importance. 
\item We introduce a question-aware attention mechanism that incorporates question context carefully in the audio-visual encoding process, enabling progressive refinement of temporal focus and more effective feature extraction. 
\item QA-TIGER achieves state-of-the-art performance across multiple benchmark datasets and provides an in-depth analysis of the impact of frame selection strategies, an area that has been previously underexplored in AVQA. 
\end{itemize}

%% file: figure/teaser.tex
\begin{figure}[ht]
    \centering
    \begin{minipage}[b]{0.48\textwidth}
        \centering
        \includegraphics[width=1.0\textwidth]{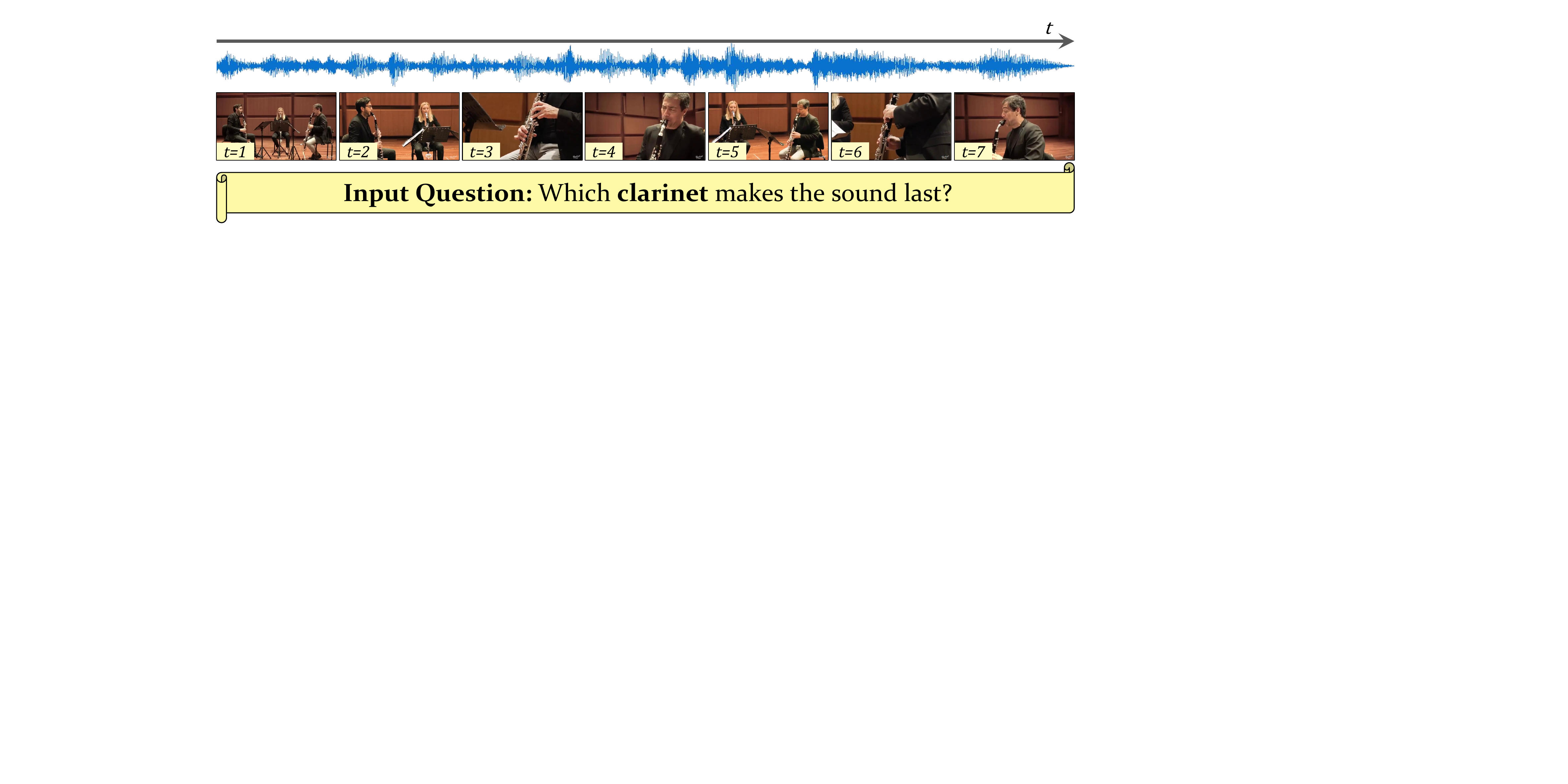}
        \vspace{-5mm}
        \subcaption{An example of input frames, audio, and question}
        \label{teaser1}
    \end{minipage}

    \begin{minipage}[b]{0.495\linewidth}
        \centering
        \includegraphics[width=1.0\textwidth]{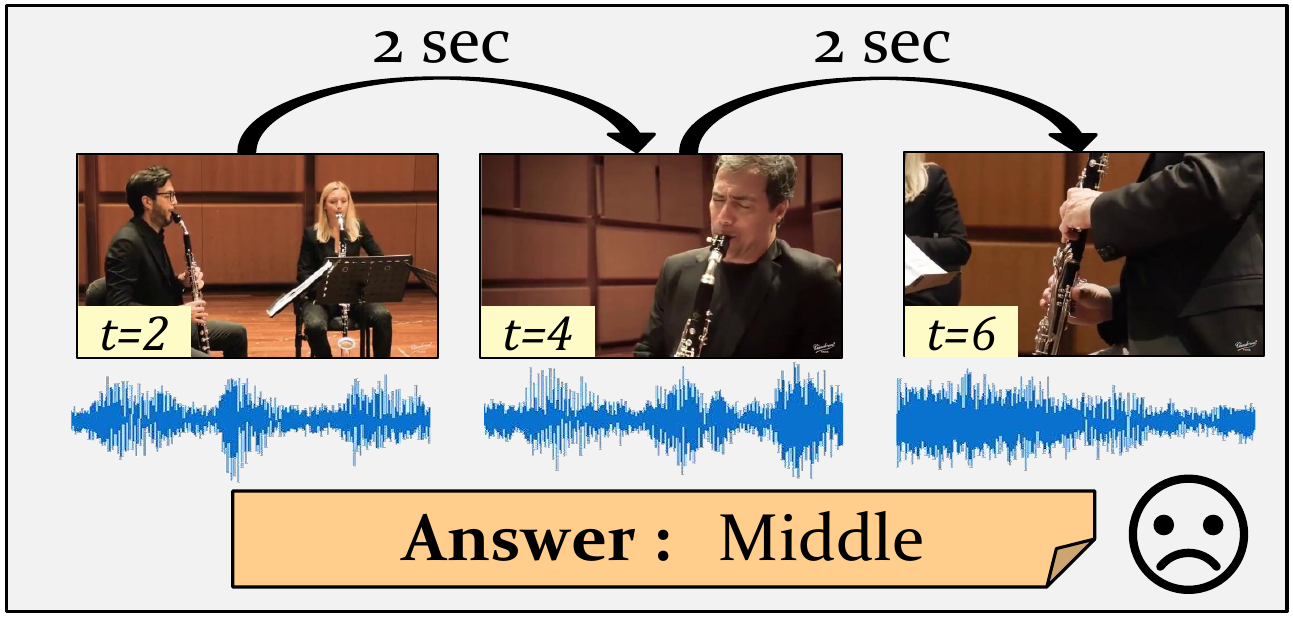}
        \subcaption{Uniform sampling} \label{teaser2}
    \end{minipage}
    \hfill %
    \begin{minipage}[b]{0.495\linewidth}
        \centering
        \includegraphics[width=1.0\textwidth]{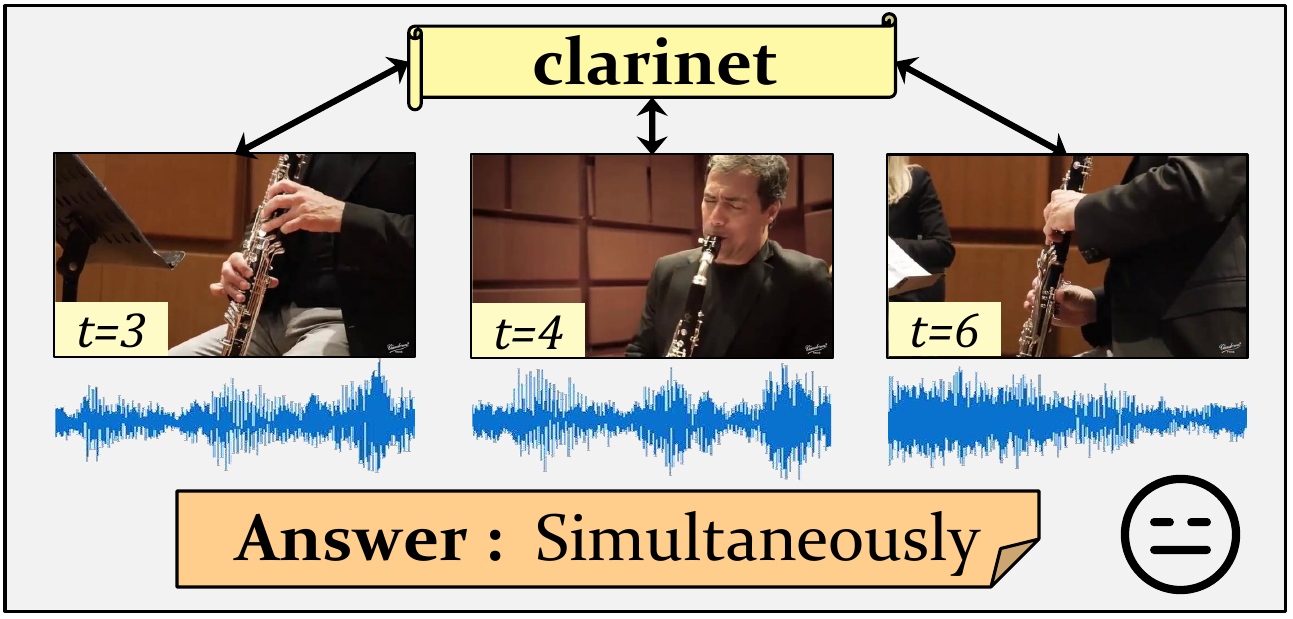}
        \subcaption{Top-K frame selection} \label{teaser3}
    \end{minipage}

    \begin{minipage}[b]{0.48\textwidth}
        \centering
        \includegraphics[width=1.0\textwidth]{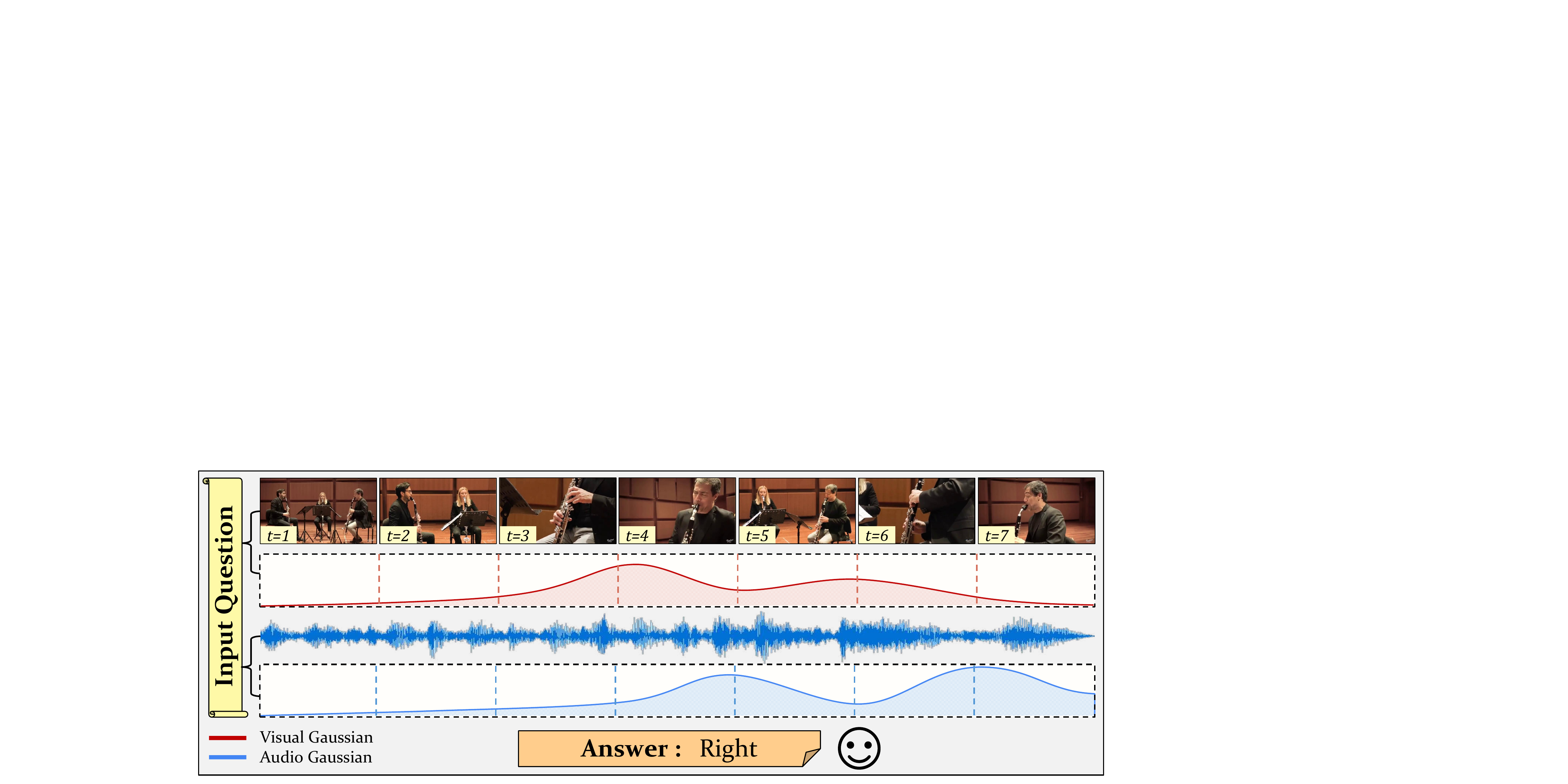}
        \subcaption{QA-TIGER} \label{teaser4}
    \end{minipage}
    \caption{Sampling comparison for AVQA. (a) Input sample. (b) Uniform sampling treats all frames equally, ignoring question-specific context. (c) Top-k frame selection samples discrete frames based only on question-video similarity, often missing important contextual details. (d) Ours provides explicit temporal grounding for both audio and visual modalities, focusing on continuous, question-relevant segments with precise multimodal alignment.}
    \label{teaser}
\end{figure}

%% file: sec/2_related_works.tex
\section{Related Works}
\label{sec:related}

\input{figure/main}

Audio-Visual Question Answering (AVQA) has advanced with datasets like MUSIC-AVQA~\cite{music-avqa22}, AVQA~\cite{avqa22}, and Pano-AVQA~\cite{yun2021pano}, enabling fine-grained reasoning~\cite{exploring22,multi23}. Early works have implemented spatial-temporal groundings for fine-grained audio-visual scene understanding and reasoning~\cite{music-avqa22, avqa22}. Recently, adapter-based architectures~\cite{lavish23, stg-cma23, dg-sct24} have achieved impressive performance by leveraging frozen pretrained vision transformers for efficiency, updating only lightweight, newly added parameters. COCA~\cite{coca23} and M2KVDG~\cite{m2k24} achieve multimodal collaboration with causal graphs, enhancing the model's robustness. APL~\cite{apl24} introduced an object-aware approach that employs adaptive-positivity learning to align question-object and audio-object semantics. Building on prior work, this study focuses on two often underexplored aspects: explicitly integrating question information and an effective frame sampling strategy to enhance temporal reasoning.

\subsection{Question Awareness}
\label{subsec:related1} 

In recent multimodal AVQA research, question-awareness~\cite{qagl23,moviechat+24,relation19,question19} has been recognized as crucial for aligning questions with visual and auditory features. 
However, most AVQA models~\cite{music-avqa22, lavish23, apl24} only incorporate question awareness at the final reasoning stage, which limits its impact throughout the entire process. 
Although several methods utilize question information for key-frame selection~\cite{pstp23,tspm24}, this is typically applied in an implicit manner, further constraining their reasoning capabilities. While models like QA-ViT~\cite{qavit24} embed question information early in the encoding process, they primarily focus on interactions between frame tokens (\,i.e., patches) and the question within a single frame. In contrast, our approach explicitly embeds question information early in the process, selectively integrating modality-relevant details along the temporal axis. This ensures consistent alignment with the question context throughout the AVQA pipeline, allowing the model to progressively refine its focus on relevant temporal cues and improving the ability to leverage both visual and audio information.

\subsection{Temporal Grounding}
\label{subsec:related2} 
Understanding and reasoning over extended audio or video, often containing redundant information, is a key challenge in QA. A critical task is localizing specific temporal segments that align with a natural language query, reducing redundancy. Temporal grounding focuses on accurately identifying these relevant moments. Many existing QA methods, such as proposal-based~\cite{semantic19, learning20, temporally20} or proposal-free~\cite{proposal21, you23, find19, mun2020local} models, like SMIN~\cite{wang2021structured}, optimize multi-level interactions between the question and video segments. However, these approaches often rely on predefined knowledge, adding computational overhead. To mitigate this, many AVQA models~\cite{mcd24, lavish23, music-avqa22, stg-cma23} use uniform sampling, which overlooks question-specific details and treats the temporal dimension as discrete. Recent studies~\cite{pstp23, tspm24} attempt to select frames based on question relevance, yet they still treat the temporal dimension as discrete, limiting their ability to capture continuous video dynamics.
In contrast, our approach embeds temporal dependencies using Gaussian distributions, enabling continuous modeling of question-relevant segments. This ensures precise alignment with both audio and visual modalities, improving temporal grounding and overall reasoning. By adapting to both consecutive and non-consecutive spans, QA-TIGER captures and utilizes relevant temporal information more effectively than methods that rely on discrete sampling.

%% file: figure/main.tex
\begin{figure*}[ht]
    \centering
    \includegraphics[width=1.0\linewidth]{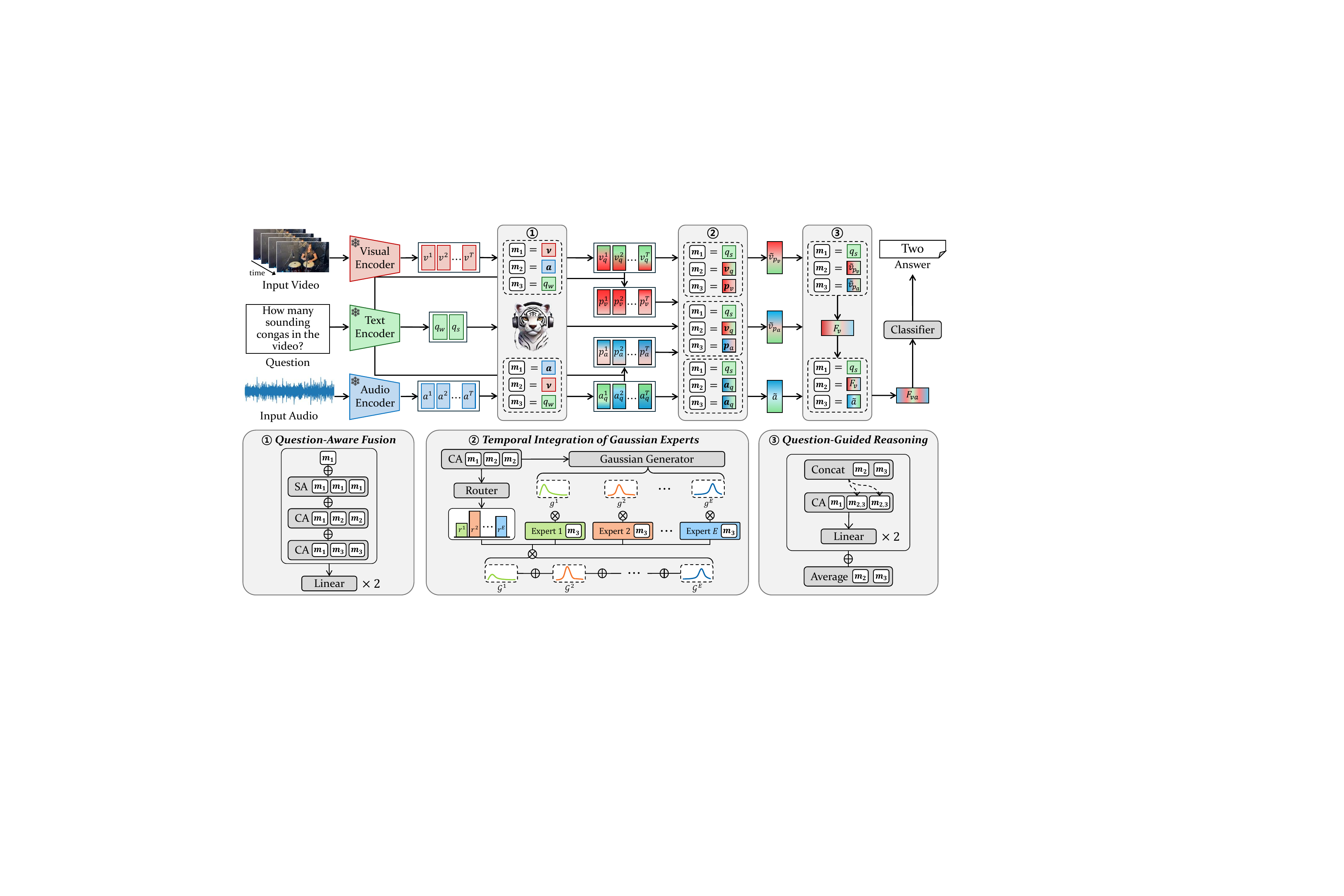}
    \vspace{-3mm}
    \caption{
    Overview of QA-TIGER: Each modality is processed through its specific pretrained encoder for feature extraction. The Question-Aware Fusion module embeds question context early in the encoding process, enabling progressive refinement throughout the pipeline. The Temporal Integration of Gaussian Experts module independently applies multi-Gaussian weighting to both audio and visual modalities, capturing consecutive and non-consecutive temporal dependencies. Finally, Question-Guided Reasoning produces context-aware audio-visual features for accurate answer prediction.
    }
    \label{Fig:main_fig}
\end{figure*}

%% file: sec/3_methods.tex
\section{Method}
\label{methods}

QA-TIGER aims to dynamically weight temporal segments based on question relevance and explicitly integrate question context for precise reasoning. Figure \ref{Fig:main_fig} illustrates the overall architecture.

\subsection{Input Representation}
The input video sequence is split into $\boldmath{T}$ non-overlapping $1s$ segments, each with paired audio and visual elements.

\noindent \textbf{Visual Representation.}
Each visual segment is processed using a pretrained CLIP model~\cite{clip21}. The visual input is divided into $M$ patches per segment, with a special [CLS] token added at the beginning. Visual features are extracted in two forms: Frame-level features $\mathbf{v} = \{v^{t}\}_{t=1}^{T} \in \mathbb{R}^{T \times D}$ are obtained from the [CLS] token output for each segment $t$. Patch-level features $\mathbf{p} = \{p^{t}\}_{t=1}^{T} \in \mathbb{R}^{T \times M' \times D}$ are obtained by merging the $M$ patch tokens into $M'$ tokens per segment using Token Merging (ToMe)~\cite{tome23}, providing more spatially detailed information.

\noindent \textbf{Audio Representation.}
For each segment $t$, we extract audio features $a^{t} \in \mathbb{R}^{D}$ using a VGGish model~\cite{vgg17} pretrained on AudioSet~\cite{audioset17}, following previous work. The complete set of audio features is $\mathbf{a} = \{a^{t}\}_{t=1}^{T} \in \mathbb{R}^{T \times D}$.

\noindent
\textbf{Question Representation.}
The input question is tokenized and processed using the CLIP text encoder. We extract sentence-level features ${q}_{s} \in \mathbb{R}^{D}$ from the [EOT] token. Additionally, we obtain word-level features ${q}_{w} \in \mathbb{R}^{N \times D}$ by skipping CLIP's final projection layer, where $N$ is the number of tokens including padding.

\subsection{Question-Aware Fusion}

For effective AVQA, our question-aware attention module explicitly injects question context into both the video and audio modalities. Our question-aware fusion module operates in two main stages. 
First, multi-head Self-Attention (\( \operatorname{SA} \)) enhances internal relationships within each modality by using the same input for the query, key, and value. Then, each modality undergoes two rounds of multi-head Cross-Attention (\( \operatorname{CA} \)): 
visual features $\mathbf{v}$ use audio features $\mathbf{a}$ as the key and value, with $\mathbf{v}$ as the query, and then apply word-level question feature $q_w$ as the key and value to align with question information. Similarly, audio features $\mathbf{a}$ use visual features $\mathbf{v}$ and question features $q_w$ sequentially as key and value, with $\mathbf{a}$ as the query. This dual CA application per modality enables explicit question-informed cross-modal interactions.
Finally, the outputs from the SA and two CA layers are combined with residual connections to enhance temporal coherence and retain alignment with the question as follows:
{
\fontsize{9}{10}
\begin{gather}
    \mathbf{v}_{q} = \mathbf{v} + \operatorname{SA}(\mathbf{v},\mathbf{v},\mathbf{v}) + \operatorname{CA}(\mathbf{v},\mathbf{a},\mathbf{a}) + \operatorname{CA}(\mathbf{v}, q_w,{q_w}), \\
    \mathbf{a}_{q} = \mathbf{a} + \operatorname{SA}(\mathbf{a},\mathbf{a},\mathbf{a}) + \operatorname{CA}(\mathbf{a},\mathbf{v},\mathbf{v}) + \operatorname{CA}(\mathbf{a},{q_w},q_w).
\end{gather}
}

The result is two question-aligned multimodal features, $\mathbf{v}_q = \{{v}_q^{t}\}_{t=1}^{T} \in \mathbb{R}^{T \times D}$ for visual and $\mathbf{a}_q = \{{a}_q^{t}\}_{t=1}^{T} \in \mathbb{R}^{T \times D}$ for audio. Unlike prior approaches that integrate question information only in the final stages, our method introduces question context from the start, allowing QA-TIGER to dynamically refine focus on relevant temporal cues throughout AVQA processing.

Once the question-aware visual features $\mathbf{v}_q$ and audio features $\mathbf{a}_q$ are obtained, we further refine the patch-level visual features to align finer spatial details with the question context. This refinement ensures that finer spatial details in the visual data align more closely with the question context embedded in the modality-specific features. The process is formulated as follows: 
\begin{gather}
    \mathbf{p}_v = \mathbf{p} + \operatorname{SA}(\mathbf{p}, \mathbf{p}, \mathbf{p}) + \operatorname{CA}(\mathbf{v}_{q}, \mathbf{p}, \mathbf{p}), \\
    \mathbf{p}_a = \mathbf{p} + \operatorname{SA}(\mathbf{p}, \mathbf{p}, \mathbf{p}) + \operatorname{CA}(\mathbf{a}_{q}, \mathbf{p}, \mathbf{p}).
\end{gather}

\subsection{Temporal Integration of Gaussian Experts}

To capture temporal dependencies in the video, we leverage the Mixture of Experts (MoE) framework~\cite{moe17}, integrating multiple Gaussian distributions across the timeline. This approach enables the model to focus on distinct temporal segments relevant to the question.

\noindent \textbf{Gaussian Generation.}
To produce question-relevant Gaussian distributions, we first generate a condensed, question-focused representation for each modality. This process aligns the visual and audio features with the question, to capture the temporal relevance of frames in both modalities. Using a cross-attention mechanism, we apply the sentence-level question feature $q_s$ to the question-aware features $\mathbf{v}_q$ and $\mathbf{a}_q$, yielding $D$-dimensional aggregated representations for each modality:
\begin{equation} \mathbf{v}'_q = \operatorname{CA}(q_s, \mathbf{v}_q, \mathbf{v}_q), \quad \mathbf{a}'_q = \operatorname{CA}(q_s, \mathbf{a}_q, \mathbf{a}_q). \end{equation}

In contrast to prior work~\cite{ngqa24}, which used a single Gaussian mask across the temporal domain, we generate multiple Gaussian distributions to capture a broader range of key frames, especially in cases where temporal variations are complex and challenging. The Gaussian distributions $\mathbf{g}={\{g^{i}\}}_{i=1}^E$ represent distinct temporal segments relevant to the question, where $E$ is the total number of Gaussian experts (\ie, Gaussian distributions):
\begin{equation} \mathbf{g}_v = \mathcal{N}(\mu^{i}_v, (\sigma^{i}_v)^2), \quad \mathbf{g}_a = \mathcal{N}(\mu^{i}_a, (\sigma^{i}_a)^2). \end{equation}

Here, $\{\mu^{i}\}_{i=1}^E$ and $\{\sigma^{i}\}_{i=1}^E$ denote the Gaussian centers and standard deviations, generated through a linear layer with an input dimension of $D$ and an output dimension of two. To minimize conflicts across time segments and enhance temporal focus, the Gaussian centers are distributed along the timeline, with a predicted offset by a linear layer added to the initial centers. This ensures that each expert mainly covers a distinct segment, aligning with question-relevant frames and reducing redundancy.

\noindent \textbf{Integrating Temporal Information.} 
To selectively capture relevant segments over time, we adapt the MoE framework, which combines multiple specialized ``experts,'' each trained to focus on specific temporal patterns. Instead of selecting discrete experts, our model adjusts the influence of each expert based on their context-dependent weights. These experts act as Gaussian distributions along the timeline, providing soft masks that help capture temporal dependencies in alignment with the question.

The router assigns routing values $\mathbf{r}_v=\{r_v^i\}_{i=1}^E$ and $\mathbf{r}_a=\{r_a^i\}_{i=1}^E$  for each expert based on the cross-attention outputs $\mathbf{v}'_q$ and $\mathbf{a}'_q$, respectively:
\begin{equation} \mathbf{r}_v = \operatorname{Softmax}(\mathbf{v}'_q \cdot W), \quad \mathbf{r}_a = \operatorname{Softmax}(\mathbf{a}'_q \cdot W),
\end{equation}
where $W \in \mathbb{R}^{D \times E}$ is a learnable weight matrix that dynamically controls the influence of each expert. The resulting expert weights allow the model to emphasize question-relevant time segments more effectively.
The outputs from all experts are then combined through a weighted summation, producing a temporally integrated representation that captures question-specific insights across frames. This integration is expressed as follows:
\begin{gather}   
    \Tilde{v}_{p_{v}} = \mathcal{G}_{v}(\mathbf{p}_{v}), \quad
    \Tilde{v}_{p_{a}} = \mathcal{G}_{v}(\mathbf{p}_{a}), \quad
    \Tilde{a} = \mathcal{G}_{a}(\mathbf{a}_q), \\
    \text{where \:\:} \mathcal{G}_{m}(x) = {\textstyle\sum_{i=1}^{E}}\, g_{m}^{i} \, r_{m}^{i} \, \mathcal{E}_{m}^{i} (x).
\end{gather}

In this setup, $\mathcal{E}_{m}^{i} (x)$ represents the output of the $i$-th expert for input $x$. The temporal visual features $\Tilde{v}_{p_{v}} \in \mathbb{R}^D$ and $\Tilde{v}_{p_{a}} \in \mathbb{R}^D$ are derived by applying the experts to modality-specific patch features $\mathbf{p}_{v}$ and $\mathbf{p}_{a}$, respectively. Similarly, the temporal audio feature $\Tilde{a} \in \mathbb{R}^D$ is obtained by applying experts to the question-aware audio feature $\mathbf{a}_q$.
This arrangement allows the model to maintain temporal coherence, aligning with question-relevant segments and minimizing redundancy.

\subsection{Question-Guided Reasoning and Prediction}

The question-guided reasoning module combines audio and visual features from the temporal integration module, allowing the model to evaluate each input’s importance in relation to the question context. To ensure a balanced representation, we use averaged features as a residual connection, preventing over-reliance on any single input type.
First, the final visual feature $F_v\in\mathbb{R}^D$  is obtained as follows:
\begin{equation}
    F_{v} = \operatorname{Avg}(\Tilde{v}_{p_a}, \Tilde{v}_{p_v}) + \operatorname{CA}(q_{s}, [\Tilde{v}_{p_a}, \Tilde{v}_{p_v}], [\Tilde{v}_{p_a}, \Tilde{v}_{p_v}]), %
\end{equation}
where [·, ·] denotes concatenation. This process yields a question-aligned visual feature that balances information across both temporal visual representations.
Next, the final audio-visual representation $F_{va} \in \mathbb{R}^D$  is obtained by fusing the temporal audio feature $\Tilde{a}$ with $F_v$ as follows:
\begin{equation} F_{va} = \operatorname{Avg}(\Tilde{a}, F_v) + \operatorname{CA}(q_{s}, [\Tilde{a}, F_v], [\Tilde{a}, F_v]). \end{equation}

This step ensures that both audio and visual features contribute to the final representation in a question-aware manner. Finally, answer prediction is performed by applying a linear layer and a Softmax layer to $F_{va}$ yielding probabilities over $C$ answer choices. The model is trained using cross-entropy loss: $\mathcal{L}_{qa} = -\sum^{C}_{c=1} y_c \log \mathcal{P}_c$, with the answer selected as the class with the highest probability.

%% file: sec/4_experiments.tex
\section{Experiments}
\label{exps}

\subsection{Datasets} 
Our experiments utilize the MUSIC-AVQA~\cite{music-avqa22}, MUSIC-AVQA-R~\cite{music-avqa-r25}, and MUSIC-AVQA-v2.0~\cite{music-avqa-v2} datasets, as summarized in Table~\ref{tab:dataset_summary}. MUSIC-AVQA serves as a benchmark for audio-visual reasoning over 22 instruments, with questions spanning audio-only, visual-only, and audio-visual modalities, covering reasoning types like existence, location, and temporal aspects. MUSIC-AVQA-R focuses on rare and out-of-distribution samples to assess model robustness. MUSIC-AVQA-v2.0 addresses dataset biases by enhancing diversity in ensemble scenes and multi-instrument cases, providing both bias and balanced sets for robust evaluation. Further details are provided in the supplementary materials.

\subsection{Implementation Details}

The video is sampled at 1 fps, with audio features extracted using VGGish~\cite{vgg17}, while visual features and questions are processed using the CLIP-ViT-L/14~\cite{clip21} model. Token reduction is applied to visual features using ToMe~\cite{tome23} technique. To ensure consistency, all features are linearly transformed to 512 dimensions. Further details on the feature extraction process are provided in the supplementary materials. We set the expert number to 7 and configure every attention module with 8 heads with 0.1 dropout probability. The model is trained with the Adam optimizer at an initial learning rate of 1e-4, which decays by a factor of 0.1 every 8 epochs. Training is conducted for 15 epochs with a batch size of 32 on a single NVIDIA RTX A6000.

\input{table/datasets}
\input{table/music-avqa}
\input{table/music-avqa-r}
\input{table/music-avqa-v2}
\input{figure/tiger_qualitative}
\subsection{Quantitative Results and Analysis}
To evaluate our model’s effectiveness, we compared QA-TIGER to existing AVQA methods on Audio (\textit{A-QA}), Visual (\textit{V-QA}), and Audio-Visual (\textit{AV-QA}) tasks, considering both question types and overall averages. QA-TIGER was trained on the MUSIC-AVQA~\cite{music-avqa22} training set and validated using its validation set for testing both MUSIC-AVQA~\cite{music-avqa22} and MUSIC-AVQA-R~\cite{music-avqa-r25}. For MUSIC-AVQA-v2.0~\cite{music-avqa-v2}, QA-TIGER was trained on both the balanced and biased training sets and tested accordingly. 

\noindent \textbf{MUSIC-AVQA.}
QA-TIGER achieves the highest performance among all models, with an overall accuracy (77.62\%), surpassing the previous best method (76.79\%), TSPM~\cite{tspm24}, in Table~\ref{tab: music-avqa}. 
Notably, our model shows strong performance in complex reasoning tasks, achieving 78.58\% and 72.50\% in \textit{AV-Counting} and \textit{AV-Local} categories respectively, surpassing previous methods.

\noindent \textbf{MUSIC-AVQA-R.}
Our method achieves 67.99\% overall accuracy on the MUSIC-AVQA-R dataset across diverse question types, as shown in Table~\ref{tab: music-avqa-r}, without the need for explicit bias handling techniques. This balanced performance highlights QA-TIGER’s strong temporal modeling and question-aware feature extraction capabilities.

\noindent \textbf{MUSIC-AVQA-v2.0.} QA-TIGER consistently outperforms existing models~\cite{music-avqa22, lavish23, music-avqa-r25} on the biased test set, regardless of the training set type (Table~\ref{tab: music-avqa-v2}a). On the balanced test set (Table~\ref{tab: music-avqa-v2}b), QA-TIGER trained on the balanced dataset achieves superior accuracy in both \textit{A-QA} (79.90\%) and \textit{V-QA} (86.95\%), surpassing LAST-Att in overall accuracy (75.44\% vs. 76.43\%). Notably, LAST-Att underperforms in \textit{A-QA} despite incorporating an additional audio encoder, Audio Spectrogram Transformer~\cite{ast21}. This highlights QA-TIGER’s robustness across diverse evaluation settings.

\noindent \textbf{Inference Time.}
We compare inference time between TSPM and QA-TIGER under identical conditions, using the full pipeline setup with CLIP and VGGish. TSPM records 1.767 seconds, while QA-TIGER achieves a comparable time of 1.737 seconds.
Despite processing full temporal information, QA-TIGER matches TSPM's efficiency—which selects Top-K frames—demonstrating its ability to handle temporal dynamics without added overhead.

\subsection{Qualitative Results of Temporal Gaussian}

To evaluate whether QA-TIGER accurately identifies question-relevant temporal segments, we qualitatively analyze its performance in Figure~\ref{Fig:qualitative}. Notably, the audio Gaussian aligns more with A-QA in Figure~\ref{qualitative1}, while the visual Gaussian is more prominent for V-QA in Figure~\ref{qualitative2}. For AV-QA, both modalities show similar Gaussian distributions in Figure~\ref{qualitative3}, demonstrating the model’s adaptive focus based on question type. A detailed comparison with uniform sampling and Top-K selection is provided in the supplementary.

\noindent \textbf{Audio Question.} QA-TIGER demonstrates its ability to handle temporal reasoning. For the question ``Is the tuba playing longer than the clarinet?", QA-TIGER’s audio Gaussian assigns high weights across the entire durations, effectively capturing the complete temporal spans essential for accurately comparing the two instruments in the question. The visual Gaussian similarly focuses on frames where the tuba and clarinet are actively played, supporting a robust comparison of their durations, as shown in Figure~\ref{qualitative1}.

\noindent \textbf{Visual Question.} Our method effectively integrates both visual and audio modalities to accurately count the saxophones, in Figure~\ref{qualitative2}. The visual Gaussian assigns higher weights to frames where all five saxophonists are clearly visible from the front, while frames with less distinct views, such as side angles or wide shots, are assigned lower weights. Meanwhile, the audio Gaussian emphasizes moments when the sounds of all five saxophones overlap, focusing on their simultaneous performance. This integration allows the model to concentrate on the most critical visual and auditory cues to predict the accurate answer.

\noindent \textbf{Audio-Visual Question.} The visual Gaussian effectively focuses on frames in the early part of the video where individual instrument appear, as well as frames where all three instruments are shown together. Correspondingly, the audio Gaussian emphasizes prominent instrument sounds. This complementary alignment enables QA-TIGER to integrate visual and auditory cues effectively, resulting in accurate identification of all instruments, as shown in Figure~\ref{qualitative3}.

\subsection{Visualization of Question-Aware Fusion}

To validate the effectiveness of our question-aware fusion module, we performed word-level visualizations in Figure~\ref{Fig:qa-qualitative}. Additional supporting examples are provided in the supplementary material. 

\noindent \textbf{``Are there saxophone and piano sound"}: In the visual modality, attention initially focuses on ``piano,'' where visual cues are subtle, then shifts to the more distinct ``saxophone.'' In the audio modality, attention consistently highlights ``piano,'' compensating for the limited visual cues, while ``saxophone," is primarily processed visually. This shows how QA-TIGER adapts to question context, effectively distributing attention across both modalities.
\input{figure/qa_qualitative}

\noindent \textbf{``How many sounding saxophone in the video"}: In the visual modality, attention focuses on ``saxophone,'' aiding in counting. In the audio modality, attention highlights ``sounding,'' and ``saxophone,'' to capture the auditory cues necessary for counting saxophone occurrences.

These results demonstrate how the question-aware fusion module dynamically adapts to different questions. By emphasizing question-relevant elements in both modalities, QA-TIGER progressively refines its focus, improving accuracy and ensuring alignment with task demands throughout the reasoning process.

%% file: table/datasets.tex
\begin{table}[tb]
\footnotesize
\centering
\begin{adjustbox}{width=\linewidth,center=\linewidth}
\begin{tabular}{l|c|c|c|c}
\bottomrule
\textbf{Dataset} & \textbf{\# Videos} & \textbf{\# Train QA} & \textbf{\# Valid QA} & \textbf{\# Test QA} \\ \hline
MUSIC-AVQA~\cite{music-avqa22}        & 9,288           & 31,904    &  4,568 & 9,129     \\
MUSIC-AVQA-R~\cite{music-avqa-r25}    & 9,288           &   -       &   -    & 211,572   \\ 
MUSIC-AVQA-v2.0~\cite{music-avqa-v2}  & 10,492          & 37,408    &  5,346 & 10,819    \\ 
\toprule
\end{tabular}
\end{adjustbox}
\vspace{-0.3cm} 
\caption{Summary of MUSIC-AVQA, MUSIC-AVQA-R, and MUSIC-AVQA-v2.0 datasets with the number of QA pairs.}
\vspace{-2.1mm} 
\label{tab:dataset_summary}
\end{table}

%% file: table/music-avqa.tex
\begin{table*}[ht]
\footnotesize
\renewcommand{\arraystretch}{1.05}
\renewcommand{\tabcolsep}{2.2mm}
\centering
\begin{adjustbox}{width=1.01\linewidth,center=\linewidth}
{
    \fontsize{7.4}{8}\selectfont
    \begin{tabular}{l|ccc|ccc|cccccc|c}
    \bottomrule
    \multirow{2}{*}{\textbf{Method}} & \multicolumn{3}{c|}{\textbf{Audio QA}} & \multicolumn{3}{c|}{\textbf{Visual QA}} & \multicolumn{6}{c|}{\textbf{Audio-Visual QA}} & \multirow{2}{*}{\textbf{Avg}} \\
                                     & \textbf{Count} & \textbf{Comp} & \textbf{Avg} & \textbf{Count} & \textbf{Local} & \textbf{Avg} & \textbf{Exist} & \textbf{Count} & \textbf{Local} & \textbf{Comp} & \textbf{Temp} & \textbf{Avg} & \\ \hline
    FCNLSTM~\cite{aqa20}                     & 70.45 & 66.22 & 68.88 & 63.89 & 46.74 & 55.21 & 82.01 & 59.34 & 46.28 & 62.15 & 47.33 & 60.06 & 60.34 \\
    BiLSTM~\cite{bilstm16}                   & 70.35 & 47.92 & 62.05 & 64.64 & 64.33 & 64.48 & 78.39 & 56.91 & 45.85 & 53.09 & 49.76 & 57.10 & 59.92 \\
    HCAttn~\cite{hcattn16}                   & 70.25 & 54.91 & 64.57 & 64.05 & 66.37 & 65.22 & 79.10 & 59.97 & 49.51 & 55.25 & 56.43 & 60.19 & 62.30 \\
    MCAN~\cite{mcan19}                       & 77.50 & 55.24 & 69.25 & 71.56 & 70.93 & 71.24 & 80.40 & 64.91 & 54.48 & 57.22 & 47.57 & 61.58 & 65.49 \\
    PSAC~\cite{psac19}                       & 75.64 & 66.06 & 72.09 & 68.64 & 69.79 & 69.22 & 77.59 & 63.42 & 55.02 & 61.17 & 59.47 & 63.52 & 66.54 \\
    HME~\cite{hme19}                         & 74.76 & 63.56 & 70.61 & 67.97 & 69.46 & 68.76 & 80.30 & 63.19 & 53.18 & 62.69 & 59.83 & 64.05 & 66.45 \\
    HCRN~\cite{hcrn20}                       & 68.59 & 50.92 & 62.05 & 64.39 & 61.81 & 63.08 & 54.47 & 53.38 & 41.53 & 52.11 & 47.69 & 50.26 & 55.73 \\
    AVSD~\cite{avsd19}                       & 72.41 & 61.90 & 68.52 & 67.39 & 74.19 & 70.83 & 81.61 & 63.89 & 58.79 & 61.52 & 61.41 & 65.49 & 67.44 \\
    Pano-AVQA~\cite{yun2021pano}             & 74.36 & 64.56 & 70.73 & 69.39 & 75.65 & 72.56 & 81.21 & 64.91 & 59.33 & 64.22 & 63.23 & 66.64 & 68.93 \\
    ST-AVQA~\cite{music-avqa22}              & 78.18 & 67.05 & 74.06 & 71.56 & 76.38 & 74.00 & 81.81 & 70.80 & 64.51 & \ul{66.01} & 63.23 & 69.54 & 71.52 \\
    COCA~\cite{coca23}                       & 79.35 & 67.68 & 75.42 & 75.10 & 75.43 & 75.23 & \textbf{83.50} & 66.63 & 69.72 & 64.12 & 65.57 & 69.96 & 72.33 \\
    PSTP-Net~\cite{pstp23}                   & 73.97 & 65.59 & 70.91 & 77.15 & 77.36 & 77.26 & 76.18 & 72.23 & 71.80 & \textbf{71.79} & 69.00 & 72.57 & 73.52 \\
    LAVISH~\cite{lavish23}                   & 82.09 & 65.56 & 75.97 & 78.98 & 81.43 & 80.22 & 81.71 & 75.51 & 66.13 & 63.77 & 67.96 & 71.26 & 74.46 \\
    APL~\cite{apl24}                         & 82.40 & \textbf{70.71} & \ul{78.09} & 76.52 & 82.74 & 79.69 & 82.99 & 73.29 & 66.68 & 64.76 & 65.95 & 70.96 & 74.53 \\
    TSPM~\cite{tspm24}                       & \ul{84.07} & 64.65 & 76.91 & \ul{82.29} & \ul{84.90} & \ul{83.61} & 82.19 & \ul{76.21} & \ul{71.85} & 65.76 & \textbf{71.17} & \ul{73.51} & \ul{76.79} \\ \hline
     \textbf{QA-TIGER} \graycell & \textbf{84.86} \graycell & \ul{67.85} \graycell & \textbf{78.58} \graycell & \textbf{83.96} \graycell & \textbf{86.29} \graycell & \textbf{85.14} \graycell& \ul{83.10} \graycell & \textbf{78.58} \graycell & \textbf{72.50} \graycell & 63.94 \graycell & \ul{69.59} \graycell & \textbf{73.74} \graycell & \bf \graycell {77.62} \\
    \toprule
    \end{tabular}
}
\end{adjustbox}
\vspace{-0.3cm}
\caption{Experimental results (\%) on the MUSIC-AVQA test set. The top-2 results are highlighted.}
\vspace{-2.2mm}
\label{tab: music-avqa}
\end{table*}    

%% file: table/music-avqa-r.tex
\begin{table*}[ht]
\renewcommand{\arraystretch}{1.05}
\renewcommand{\tabcolsep}{1.2mm}
\centering
\begin{adjustbox}{width=\linewidth,center=\linewidth}
\begin{tabular}{l|cccc|cccc|cccccccccc|c}
\bottomrule
\multirow{3}{*}{\textbf{Method}} & \multicolumn{4}{c|}{\textbf{Audio QA}} & \multicolumn{4}{c|}{\textbf{Visual QA}} & \multicolumn{10}{c|}{\textbf{Audio-Visual QA}} & \multirow{3}{*}{\textbf{Avg}} \\
 & \multicolumn{2}{c}{\textbf{Count}} & \multicolumn{2}{c|}{\textbf{Comp}} & \multicolumn{2}{c}{\textbf{Count}} & \multicolumn{2}{c|}{\textbf{Local}} & \multicolumn{2}{c}{\textbf{Exist}} & \multicolumn{2}{c}{\textbf{Count}} & \multicolumn{2}{c}{\textbf{Local}} & \multicolumn{2}{c}{\textbf{Comp}} & \multicolumn{2}{c|}{\textbf{Temp}} & \\
\cline{2-19}
 & H & T & H & T & H & T & H & T & H & T & H & T & H & T & H & T & H & T & \\
\hline
FCNLSTM~\cite{aqa20} & 66.23 & 36.48 & 64.78 & 51.24 & 61.75 & 5.31 & 54.86 & 51.06 & 64.76 & 78.52 & 62.69 & 7.23 & \ul{46.66} & 57.30 & 43.13 & 71.67 & 37.02 & 30.78 & 54.12 \\
BiLSTM~\cite{bilstm16} & 73.68 & 46.32 & 21.51 & \textbf{77.58} & 64.30 & 0.00 & 53.92 & 42.01 & \textbf{87.51} & 21.14 & 62.85 & 2.18 & 35.16 & 43.75 & 27.61 & \ul{74.38} & 17.58 & 31.32 & 48.84 \\
HCAttn~\cite{hcattn16} & 61.67 & 41.63 & 59.09 & 47.14 & 56.52 & 9.20 & 67.01 & 53.16 & 66.57 & 61.13 & 59.53 & 12.48 & 37.05 & 42.48 & 48.81 & 60.12 & 33.82 & 39.26 & 51.90 \\
MCAN~\cite{mcan19} & 75.02 & 60.16 & 58.89 & 50.09 & 64.58 & 26.69 & 66.48 & 62.25 & 51.29 & 67.29 & 64.76 & 25.28 & 46.11 & 61.61 & 50.57 & 52.40 & 34.64 & \ul{58.05} & 57.27 \\
PSAC~\cite{psac19} & 53.01 & 56.68 & 57.41 & 48.12 & 49.55 & 26.43 & 72.96 & 60.69 & 50.56 & 55.54 & 56.70 & 19.58 & 41.98 & 52.30 & 38.13 & 58.92 & 26.68 & 46.24 & 50.45 \\
HME~\cite{hme19} & 62.60 & 53.95 & 54.97 & 58.29 & 50.95 & 16.46 & 73.25 & 58.60 & 65.74 & 66.49 & 63.18 & 17.18 & 33.79 & 46.03 & 53.20 & 69.57 & 33.95 & 41.57 & 53.66 \\
HCRN~\cite{hcrn20} & 55.53 & 53.31 & 47.17 & 32.44 & 41.87 & 23.55 & 39.40 & 51.27 & 41.81 & 65.45 & 54.58 & 19.57 & 36.62 & 42.72 & 33.33 & 36.87 & \ul{40.47} & 44.13 & 43.92 \\
AVSD~\cite{avsd19} & 54.00 & 47.84 & 60.61 & 47.79 & 60.34 & 10.07 & 74.78 & 61.43 & 66.28 & 61.98 & 46.21 & 8.06 & 33.00 & 40.35 & 51.98 & 66.00 & 40.14 & 41.52 & 52.33 \\
Pano-AVQA~\cite{yun2021pano} & 50.57 & 43.45 & 50.78 & 44.93 & 47.28 & 15.50 & 67.19 & 65.51 & 52.37 & 22.04 & 52.21 & 21.52 & 44.35 & \ul{61.69} & 45.61 & 40.49 & 35.00 & 49.33 & 47.40 \\
ST-AVQA~\cite{music-avqa22} & 56.40 & 41.48 & 62.28 & 57.59 & 59.86 & 12.94 & 63.31 & 54.00 & 73.35 & 77.26 & 48.31 & 8.41 & 35.35 & 40.49 & \ul{53.30} & 62.44 & 40.25 & 38.15 & 52.80 \\
LAVISH~\cite{lavish23} & 61.73 & 43.99 & 65.06 & \ul{60.38} & 65.53 & 11.13 & 70.21 & 64.73 & \ul{77.83} & \ul{79.46} & 49.88 & 14.87 & 41.76 & 41.20 & \textbf{59.26} & 65.10 & \textbf{41.84} & 46.26 & 57.63 \\
TSPM$\dagger$~\cite{tspm24} & \ul{81.65} & \ul{71.80} & \ul{67.66} & 49.56 & \ul{78.29} & \ul{47.56} & \ul{80.58} & \ul{73.18} & 69.15 & \textbf{82.79} & \textbf{77.09} & \textbf{38.64} & 42.24 & 57.37 & 52.07 & 68.86 & 39.23 & 49.36 & \ul{66.30} \\ \hline
\textbf{QA-TIGER} \graycell & \textbf{82.67} \graycell & \textbf{75.82} \graycell & \textbf{71.75} \graycell & 43.11 \graycell & \textbf{81.30} \graycell & \textbf{54.59} \graycell & \textbf{84.76} \graycell & \textbf{75.59} \graycell & 72.84 \graycell & 78.56 \graycell & \ul{76.70} \graycell & \ul{33.55} \graycell & \textbf{48.22} \graycell & \textbf{64.65} \graycell & 37.55 \graycell & \textbf{80.47} \graycell & 36.85 \graycell & \textbf{62.96} \graycell & \bf \graycell {67.99} \\
\toprule
\end{tabular}
\end{adjustbox}
\vspace{-4.5mm}
\begin{flushleft}
\end{flushleft}
\vspace{-0.7cm}
\caption{Experimental results (\%) on the MUSIC-AVQA-R test set, with H and T representing performance on Head (frequent) and Tail (rare) answer categories, respectively. $\dagger$ indicates results reproduced using official codes.}
\vspace{-2.2mm}
\label{tab: music-avqa-r}
\end{table*}

%% file: table/music-avqa-v2.tex
\begin{table}[tb]
\renewcommand{\arraystretch}{1.05}
\renewcommand{\tabcolsep}{1.2mm}
\footnotesize
\centering
\begin{adjustbox}{width=\linewidth,center=\linewidth}
\begin{tabular}{c|c|l|cccc}
\bottomrule
\textbf{Test} & \textbf{Training} & \textbf{Method} & \textbf{A-QA} & \textbf{V-QA} & \textbf{AV-QA} & \textbf{Avg} \\
\hline
\multirow{7}{*}{\small (a) Bias} 
                                            & \multirow{3}{*}{Bias}        & ST-AVQA~\cite{music-avqa22}      & \ul{76.86}     & 77.70          & 69.59          & 73.07          \\
                                            &                              & LAVISH~\cite{lavish23}           & 76.73          & \ul{80.96}     & \ul{70.80}     & \ul{74.59}     \\
                                            &                              & \textbf{QA-TIGER}                & \textbf{79.13} & \textbf{84.83} & \textbf{72.37} & \textbf{76.93} \\ \cline{2-7}
                                            & \multirow{4}{*}{Balance}     & ST-AVQA~\cite{music-avqa22}      & 76.18          & 77.20          & 67.96          & 71.92          \\
                                            &                              & LAVISH~\cite{lavish23}           & 75.56          & 80.83          & 69.27          & 73.51          \\
                                            &                              & LAST~\cite{music-avqa-v2}        & \ul{77.10}     & 82.99          & 70.86          & 75.24          \\
                                            &                              & LAST-Att~\cite{music-avqa-v2}    & \textbf{77.29} & \ul{83.47}     & \ul{71.05}     & \ul{75.45}     \\
                                            &                              & \textbf{QA-TIGER}                & 77.07          & \textbf{85.93} & \textbf{71.20} & \textbf{76.57} \\
\toprule
\addlinespace[1.0em]
\bottomrule
\textbf{Test} & \textbf{Training} & \textbf{Method} & \textbf{A-QA} & \textbf{V-QA} & \textbf{AV-QA} & \textbf{Avg} \\
\hline
\multirow{7}{*}{\small (b) Balance} 
                                            & \multirow{3}{*}{Bias}        & ST-AVQA~\cite{music-avqa22}   & \ul{73.34}     & 76.82          & 64.51          & 69.40          \\
                                            &                              & LAVISH~\cite{lavish23}        & 73.14          & \ul{79.70}     & \ul{65.01}     & \ul{70.39}     \\
                                            &                              & \textbf{QA-TIGER}             & \textbf{77.57} & \textbf{84.84} & \textbf{67.43} & \textbf{73.91} \\ \cline{2-7}
                                            & \multirow{4}{*}{Balance}     & ST-AVQA~\cite{music-avqa22}   & 75.50          & 77.67          & 66.32          & 71.02          \\
                                            &                              & LAVISH~\cite{lavish23}        & 76.15          & 81.32          & 68.28          & 73.18          \\
                                            &                              & LAST~\cite{music-avqa-v2}     & 78.08          & 83.29          & 69.72          & 74.85          \\
                                            &                              & LAST-Att~\cite{music-avqa-v2} & \ul{78.56}     & \ul{84.07}     & \textbf{70.30} & \ul{75.44}     \\
                                            &                              & \textbf{QA-TIGER}             & \textbf{79.90} & \textbf{86.95} & \ul{70.22}     & \textbf{76.43} \\
\toprule
\end{tabular}
\end{adjustbox}
\caption{Experimental results (\%) on the MUSIC-AVQA-v2.0 for (a) bias and (b) balanced test sets. Each evaluation is conducted with models trained on bias and balanced training sets, respectively. The performance of other methods was referenced from the scores provided by MUSIC-AVQA-v2.0.}
\label{tab: music-avqa-v2}
\vspace{-10mm}
\end{table}

%% file: figure/tiger_qualitative.tex
\begin{figure*}[t]
    \centering
    \begin{minipage}[b]{1.0\textwidth}
        \centering
        \includegraphics[width=1.0\textwidth]{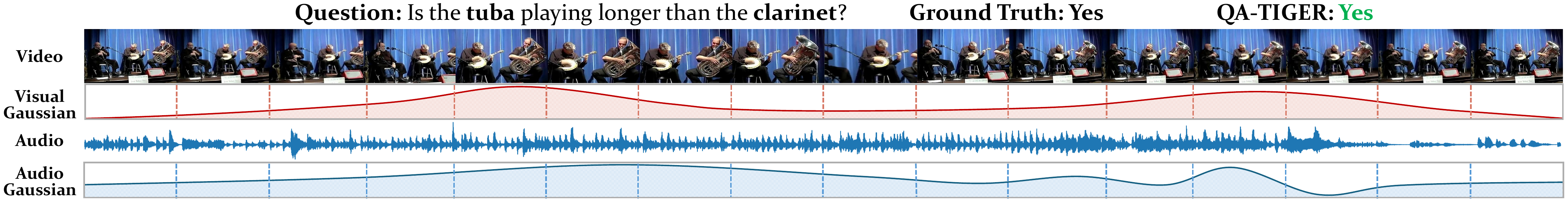}
        \subcaption{Audio Question}
        \label{qualitative1}
    \end{minipage}

    \begin{minipage}[b]{1.0\textwidth}
        \centering
        \includegraphics[width=1.0\textwidth]{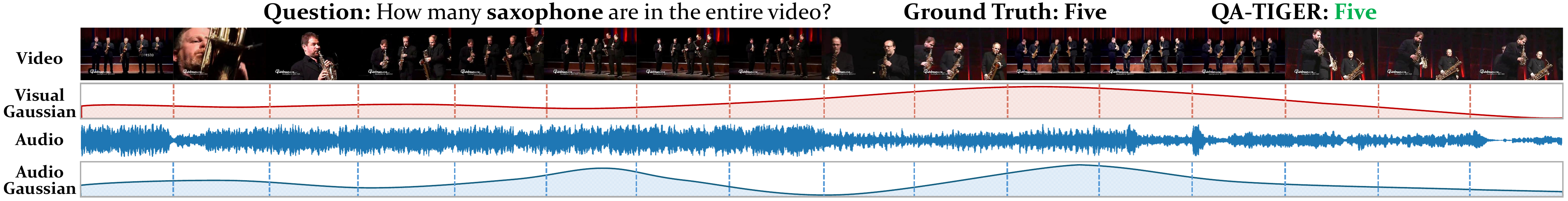}
        \subcaption{Visual Question}
        \label{qualitative2}
    \end{minipage}

    \begin{minipage}[b]{1.0\textwidth}
        \centering
        \includegraphics[width=1.0\textwidth]{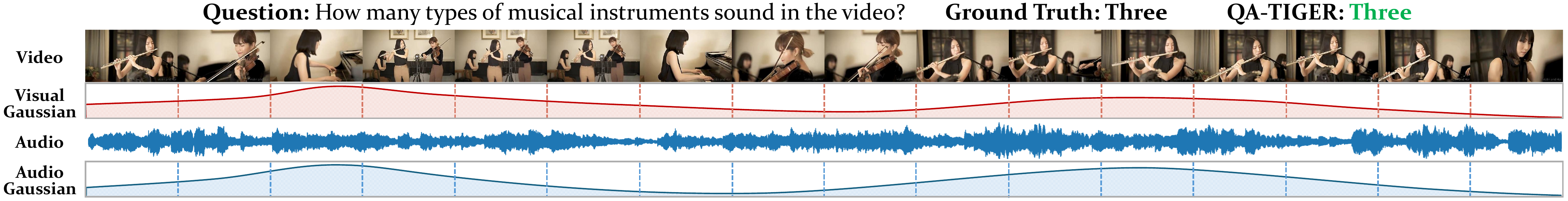}
        \subcaption{Audio-Visual Question}
        \label{qualitative3}
    \end{minipage}
    \vspace{-5mm}
    \caption{Qualitative results by question type of the proposed QA-TIGER on the MUSIC-AVQA test set.}
    \label{Fig:qualitative}
    \vspace{-2mm}
\end{figure*}

%% file: figure/qa_qualitative.tex
\begin{figure}[t]
    \centering
    \includegraphics[width=1.0\linewidth]{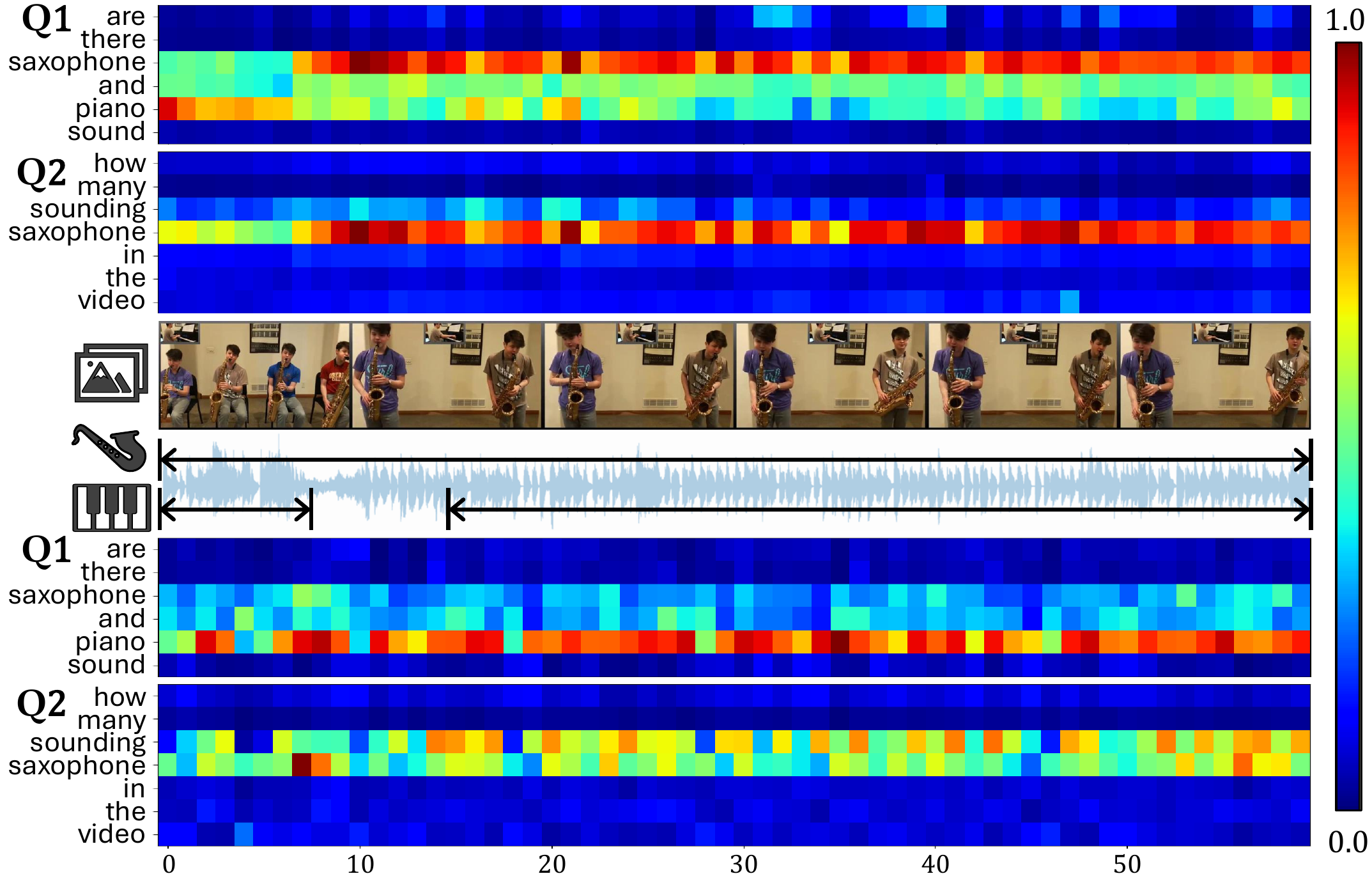}
    \vspace{-4mm}
    \caption{Question-aware attention for visual (upper) and audio (lower) fusion in response to two different questions for the same video and audio input. The attention intensity is indicated by the color scale on the right.}
    \label{Fig:qa-qualitative}
    \vspace{-6mm}
\end{figure}

%% file: sec/5_ablation.tex
\input{table/module_ablation}
\section{In-depth Analysis}
\label{ablation}

\subsection{Ablation Study}
We evaluated our proposed modules on the MUSIC-AVQA dataset (Table~\ref{tab: module_ablation}). The baseline model used uniform sampling and achieved reasonable accuracy. Adding the \textit{Gaussian Experts} module improved performance by enhancing temporal modeling through specialized experts. The \textit{Question-Aware Fusion} module alone also contributed a notable improvement. Combining both modules led to the best performance, demonstrating the synergistic benefits of integrating question context early on. This integration enriches the temporal context for the \textit{Gaussian Experts} module, highlighting the complementary strengths of both modules in enhancing temporal reasoning.
Without \textit{Question-Aware Fusion}, question features were incorporated only at the final stage, as in prior work~\cite{music-avqa22,lavish23,AVMamba24}. For setups excluding \textit{Gaussian Experts}, conventional uniform sampling was applied.

\subsection{Frame Sampling}

We evaluated conventional sampling strategies and Gaussian modeling designs for AVQA tasks (Figure~\ref{Fig:method_comparison}). 
Note that all experiments compare only the frame sampling part after applying the proposed \textit{Question-Aware Fusion}.

\noindent \textbf{Sampling-based Methods.} \textit{Uniform} sampling at fixed intervals was less effective due to its lack of focus on question-relevant frames. The \textit{Top-K} method, using the frame selection module from TSPM~\cite{tspm24}, improved performance by selecting question-relevant frames but lacked comprehensive temporal context.

\noindent \textbf{Gaussian-based Methods.} We explore various Gaussian-based methods to model temporal dependencies. \textit{Gaussian}, a simple sum of Gaussians, shows similar performance to uniform sampling. \textit{Weighted-Gaussian} introduces dynamic weighting but still struggles with overlap. By adding disjoint center constraints, \textit{Weighted-Gaussian (Disjoint Center)} reduces redundancy and improves accuracy. Building on this, \textit{Gaussian Experts} extends the \textit{Weighted-Gaussian (Disjoint Center)} approach with a MoE framework, dynamically assigning specialized experts to distinct temporal segments, resulting in the highest accuracy.

\subsection{Selection of Experts}

In this analysis, we explored the impact of expert selection in the proposed MoE approach. The decision to activate all seven experts stems from the observed performance gains as more experts are used. Even with the fewest experts, the model outperforms the current state-of-the-art TSPM~\cite{tspm24} (76.79\%) on the MUSIC-AVQA~\cite{music-avqa22} dataset, as shown in Figure~\ref{Fig:expert_selection}. Using all seven experts achieves the highest accuracy (77.62\%), showcasing the superiority of QA-TIGER.

\input{figure/selection_comparison}
\input{figure/expert_selection}

%% file: table/module_ablation.tex
\begin{table}[h]
\centering
\begin{adjustbox}{width=\linewidth,center=\linewidth}
\begin{tabular}{c|c|cccc}
\bottomrule
\multicolumn{1}{c|}{\textbf{\begin{tabular}[c]{@{}c@{}}Question-Aware\\ Fusion\end{tabular}}} & \multicolumn{1}{c|}{\textbf{\begin{tabular}[c]{@{}c@{}}Gaussian\\ Experts\end{tabular}}} & \textbf{A-QA}  & \textbf{V-QA}  & \textbf{AV-QA} & \textbf{Avg}  \\ \hline
                                                      &                                                       & 75.05          & 81.79          & 71.82          & 75.04          \\ 
\checkmark                                            &                                                       & 77.59          & 84.56          & 72.37          & 76.53          \\
                                                      &  \checkmark                                           & 76.35          & 82.74          & 72.78          & 76.05          \\
\checkmark                                            &  \checkmark                                           & \textbf{78.58} & \textbf{85.14} & \textbf{73.74} & \textbf{77.62} \\  
\toprule
\end{tabular}
\end{adjustbox}
\vspace{-0.1cm}
\caption{Ablation study of the proposed framework.}
\vspace{-2mm}
\label{tab: module_ablation}
\end{table}

%% file: figure/selection_comparison.tex
\begin{figure}[tb]
    \centering
    \includegraphics[width=0.92\linewidth]{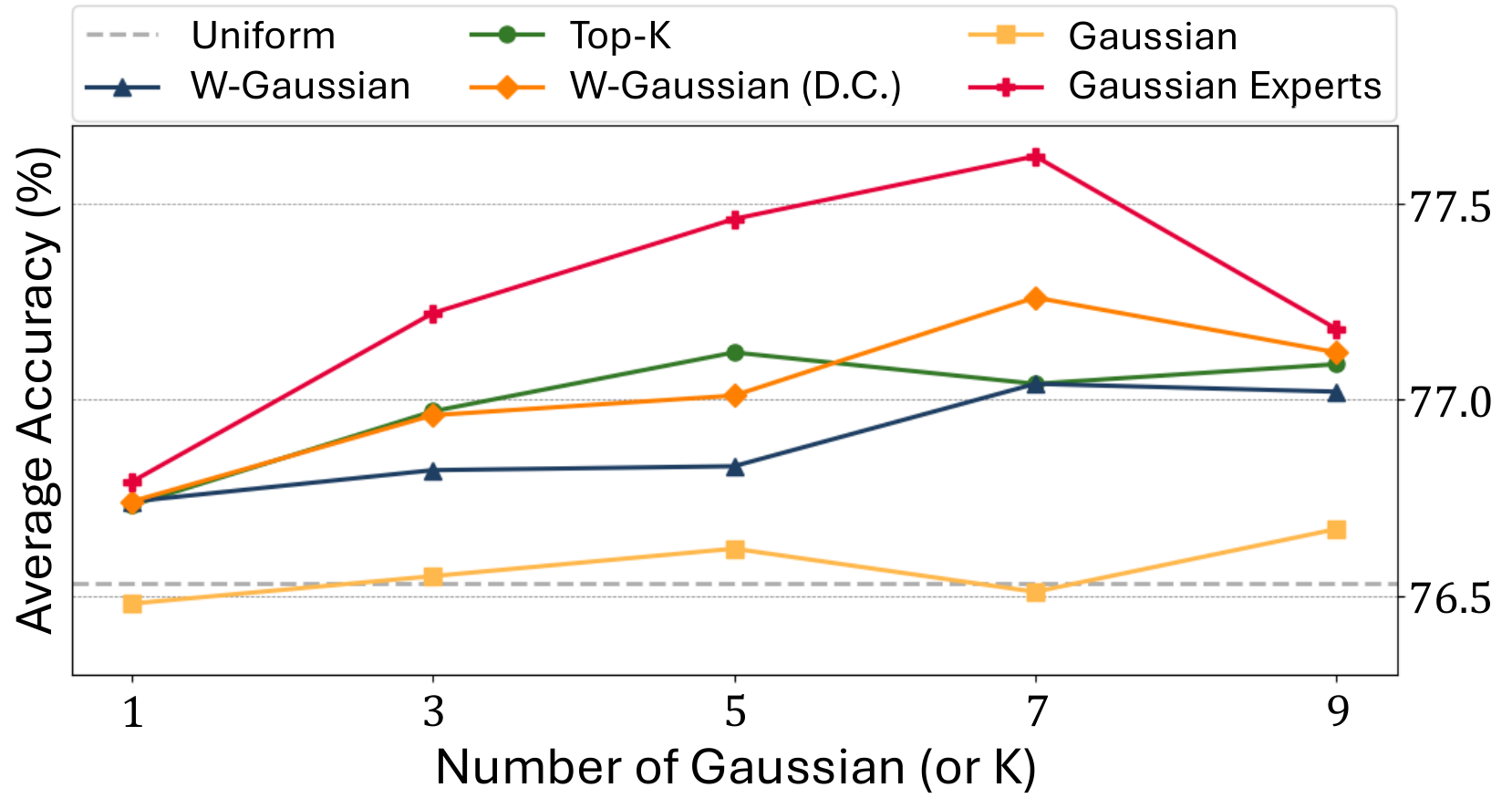}
    \caption{Comparison of temporal selection methods based on the number of criteria (Gaussian or K). Gaussian-based methods use a simple summation for temporal aggregation. ``W'' refers to ``Weighted sum,'' and ``D.C.'' stands for ``Disjoint Center.''}
    \label{Fig:method_comparison}
    \vspace{-2mm}
\end{figure}

%% file: figure/expert_selection.tex
\begin{figure}[tb]
    \centering
    \includegraphics[width=0.75\linewidth]{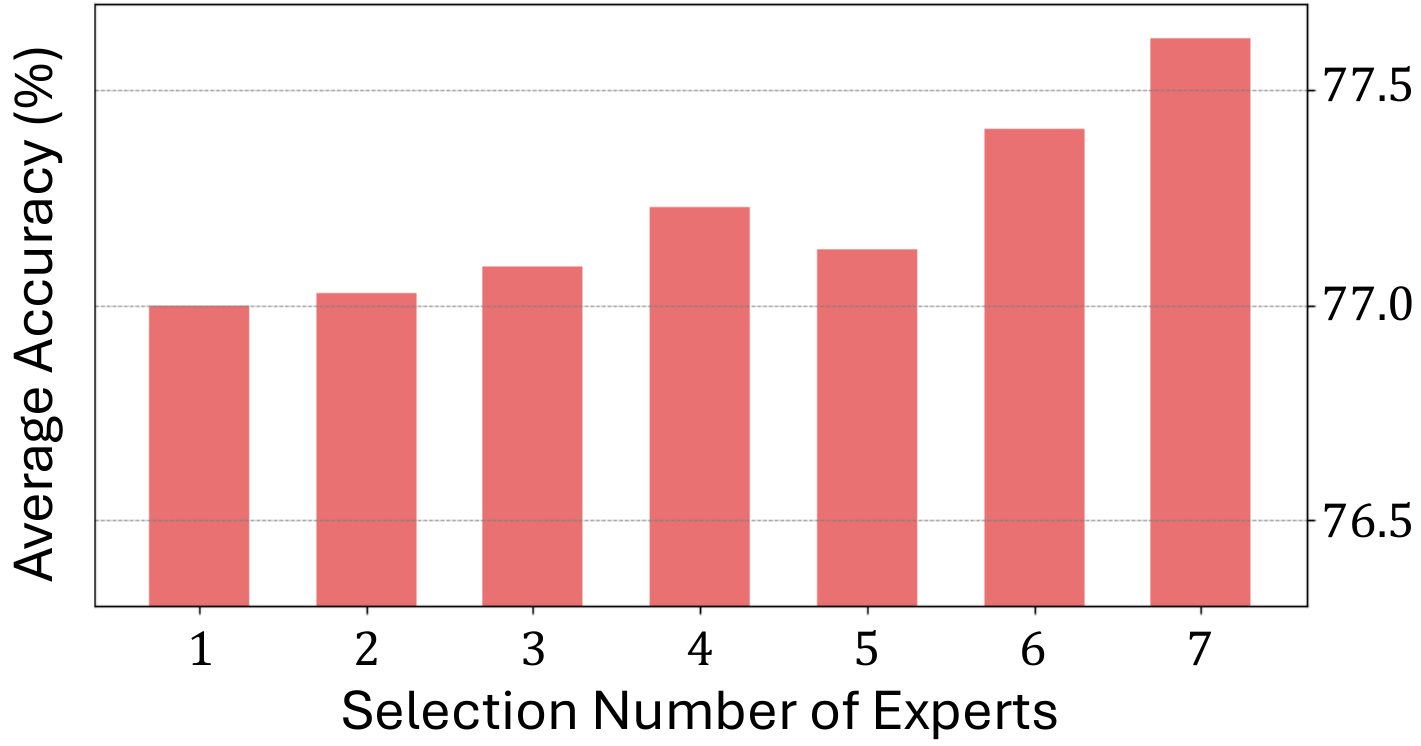}
    \vspace{-1mm}
    \caption{Impact of expert selection on Gaussian experts.}
    \label{Fig:expert_selection}
    \vspace{-2mm}
\end{figure}

%% file: sec/6_conclusion.tex
\section{Conclusion}
\label{conclusion}

We have introduced QA-TIGER, designed to overcome the limitations of prior methods in handling complex temporal dynamics and question-specific feature integration. By leveraging multi-Gaussian modeling within the MoE framework, QA-TIGER adaptively captures subtle temporal dependencies while minimizing redundancy through strategically positioned Gaussian centers. We also incorporate question context explicitly from the encoding stages, ensuring that question information is gradually refined throughout the processing pipeline. Extensive experiments demonstrate QA-TIGER’s state-of-the-art performance on various datasets, highlighting its effectiveness in tackling the continuous temporal domain and complex AVQA challenges. Future work could explore adaptive expert selection mechanisms in MoE-based AVQA, tailored to different question types, to further enhance the model's generalizability.

%% file: sec/X_suppl.tex
\appendix
\renewcommand\thefigure{\Alph{figure}}    
\setcounter{figure}{0}  
\renewcommand\thetable{\Alph{table}}
\setcounter{table}{0} 
\maketitlesupplementary

\section{Experimental Setup}
\subsection{Dataset}
\noindent \textbf{MUSIC-AVQA.\footnote{\scriptsize\hypersetup{urlcolor=magenta}\url{https://github.com/GeWu-Lab/MUSIC-AVQA}}}
We use the MUSIC-AVQA dataset~\cite{music-avqa22} to train and test our model. This dataset is a comprehensive benchmark specifically designed for audio-visual reasoning tasks. It contains 9,288 music performance videos sourced from YouTube, totaling over 150 hours of content. The dataset features 22 different instruments and includes 45,869 question-answer pairs.
The questions are categorized into audio-only, visual-only, and audio-visual types based on 33 templates. These templates cover a range of reasoning categories, such as existence, location, quantity, comparison, and temporal aspects. MUSIC-AVQA excels in challenging models with complex spatio-temporal reasoning, making it a crucial benchmark for evaluating advanced audio-visual understanding.

\vspace{+1mm} \noindent \textbf{MUSIC-AVQA-R.\footnote{\scriptsize\hypersetup{urlcolor=magenta}\url{https://github.com/reml-group/MUSIC-AVQA-R}}}
We also evaluate our model's performance on MUSIC-AVQA-R~\cite{music-avqa-r25}, an extended version of the original MUSIC-AVQA dataset designed to test the model's robustness. This extension restructures and significantly expands the test set, with a particular emphasis on rare cases and out-of-distribution samples. It includes 211,572 restructured questions, offering a more comprehensive evaluation across a wider range of question types beyond basic template-based questions. This makes MUSIC-AVQA-R a rigorous benchmark for assessing models' capabilities in spatial-temporal reasoning and their ability to manage complex multimodal interactions.

\vspace{+1mm} \noindent  \textbf{MUSIC-AVQA-v2.0.\footnote{\scriptsize\hypersetup{urlcolor=magenta}\url{https://github.com/DragonLiu1995/MUSIC-AVQA-v2.0}}}
We assess our model on MUSIC-AVQA-v2.0~\cite{music-avqa-v2}, an improved version of the original MUSIC-AVQA dataset that addresses data bias issues. This updated dataset offers a more balanced benchmark for audio-visual question answering, containing 53,573 question-answer pairs with a broader range of musical ensembles and more complex audio-visual interactions. To reduce bias, the authors manually curated 1,230 additional musical performance videos from YouTube and created 8,100 new QA pairs to supplement the original dataset. These updates ensure more balanced answers across various question templates.

\subsection{Implementation Details}
\noindent \textbf{Disjoint-Centered Gaussian Experts.}
To reduce overlaps in the regions of influence among multiple Gaussian distributions, we initialize the center positions $\mathbf{u}_{\text{fixed}}$ of the $E$ Gaussian experts with a predefined margin between them (Algorithm~\ref{algo:gaussian_experts}). While minor overlap may occur due to the Gaussian widths, the centers remain non-overlapping, reducing redundant temporal influences. 
To refine temporal segments, learnable offsets adjust the fixed positions, keeping centers within constrained margins. This ensures each expert focuses on distinct temporal ranges, capturing question-relevant segments more effectively while minimizing redundancy and maintaining expert specialization.
\input{algorithm/gaussian_experts}

\vspace{+1mm}
\noindent \textbf{Patch Merging.}
To ensure a fair comparison with TSPM~\cite{tspm24}, we utilize the patch merging strategy from ToMe~\cite{tome23}, which merges similar visual tokens within each transformer block. This involves dividing tokens into subsets, calculating similarities, and applying mean fusion to generate merged token features. By adopting this method, we align the feature extraction pipeline with TSPM for consistency in the experimental setup. For more detail on patch merging, please refer to ToMe\footnote{\scriptsize\hypersetup{urlcolor=magenta}\url{https://github.com/facebookresearch/ToMe}} and TSPM\footnote{\scriptsize\hypersetup{urlcolor=magenta}\url{https://github.com/GeWu-Lab/TSPM}}.

\input{table/encoder_comp}
\section{More Experimental Results}
Since some prior studies evaluate the model with CLIP-B/32 encoder, we evaluate our method under the same setting to enable direct comparison. Previous studies, such as PSTP-Net~\cite{pstp23} and TSPM~\cite{tspm24}, have shown that performance improves when transitioning from CLIP-B/32 to a more advanced feature extractor like CLIP-L/14. However, using CLIP-B/32 remains a relevant benchmark for assessing the baseline performance of methods. As shown in Table~\ref{tab:encoder_comp}, our method consistently outperforms PSTP-Net and TSPM, even with the smaller CLIP-B/32 encoder. This demonstrates the robustness of our approach, highlighting that its effectiveness is not solely reliant on high-capacity encoders.

\input{figure/visual_gaussian_example}

\input{figure/supple_pos_a_attention}

\input{figure/supple_pos_v_attention}

\input{figure/supple_pos_av_attention}

\section{In-Depth Visualization of QA-TIGER}
We demonstrate how QA-TIGER dynamically integrates temporal and multimodal information through Gaussian experts and question-aware fusion across diverse scenarios.

\subsection{Temporal Integration of Gaussian Experts}
Gaussian experts adaptively focus on specific temporal regions based on the input and question.
Figure~\ref{fig:gaussian_example} illustrates the temporal weights generated by the seven Gaussian experts employed in the model (see Figures~\ref{visual_experts} and~\ref{audio_experts}). These graphs show how the weights are combined to emphasize the model's focus on distinct temporal segments across visual and audio modalities (refer to Figures~\ref{visual_gaussian} and~\ref{audio_gaussian}). Each Gaussian curve demonstrates that the experts specialize in specific, minimally overlapping temporal regions, dynamically adjusting their focus based on the question and modality. By integrating these Gaussian weights, the model selectively attends to different frames according to the assigned expert and modality, even when processing the same question. This frame-level and modality-specific information is then utilized in the question-guided reasoning stage.

\subsection{Visualization of Question-Aware Fusion}
We visualize the attention of the question-aware fusion module across nine question types to highlight the effectiveness of our method. These nine question types are categorized into three main groups: Audio-QA (A-QA), Visual-QA (V-QA), and Audio-Visual-QA (AV-QA). A-QA focuses solely on auditory cues, V-QA relies exclusively on visual information, while AV-QA requires the integration of both modalities to address multi-modal questions effectively. Note that attention is presented in two parts: visual modality (top) and audio modality (bottom).

\input{figure/suppl_neg_attention}

\subsubsection{Valid Cases}
\vspace{+1mm} \noindent \textbf{\textit{Audio} Counting.}
In the visual modality, attention focuses on the ``bassoon'' with two players clearly visible, while the ``clarinet'' is partially obscured (Figure~\ref{fig:audio_counting}). The audio modality highlights the ``clarinet'', ensuring recognition of both instruments. Interestingly, the word ``are'' also gains attention, likely due to its role in framing the question involving counting and the presence of instruments.

\vspace{+1mm} \noindent \textbf{\textit{Audio} Comparative.}
The attention mechanism dynamically shifts between modalities to adapt to changing contexts. Initially, the audio attention focuses on the ``piano'' due to its prominent sound. As the ``flute'' becomes more dominant, the visual attention compensates by identifying that the ``piano'' continues to be played, even though its audio presence has diminished. This interaction is illustrated in Figure~\ref{fig:audio_comparison}.

\vspace{+1mm} \noindent \textbf{\textit{Visual} Counting.}
In the visual modality, attention focuses on ``many'' and ``saxophone,'' leveraging visual cues to estimate the number of instruments. Meanwhile, the audio modality complements this by highlighting the distinct auditory features of saxophone sounds, helping to identify and differentiate instances throughout the video in Figure~\ref{fig:visual_counting}.

\vspace{+1mm} \noindent \textbf{\textit{Visual} Location.}
The fusion module focuses on ``kind of'' and ``leftest instrument'' in the visual modality, using spatial cues to locate the leftmost instrument. In contrast, the audio modality emphasizes ``kind of'' and ``instrument'' to classify its type. Together, these modalities effectively balance spatial and categorical aspects as illustrated in Figure~\ref{fig:visual_location}.

\vspace{+1mm} \noindent \textbf{\textit{Audio-Visual} Existential.}
When observing the entire video from a distance, the visual modality primarily focuses on the ``bagpipe'', determining whether it is consistently visible throughout the scene. Meanwhile, the audio modality emphasizes the words ``always'' and ``playing'', assessing whether the bagpipe consistently produces sound. This interplay between the modalities is illustrated in Figure~\ref{fig:av_exist}.

\vspace{+1mm} \noindent \textbf{\textit{Audio-Visual} Counting.}
The fusion module adapts its attention to focus on counting-related cues across modalities. In the visual modality, attention emphasizes ``drum'' and the word ``many'' through close-up images, enabling accurate counting of the drums. At the same time, it highlights ``sounding'' and ``drum'' in the audio modality to distinguish individual drum sounds as depicted in Figure~\ref{fig:av_count}.

\vspace{+1mm} \noindent \textbf{\textit{Audio-Visual} Location.}
In this synthetic video, with only the flute sound present, the model uses spatial and auditory cues. Visually, it focuses on ``left'' and ``sounding'' to locate the instrument, while auditorily, it emphasizes ``instrument'' to classify its type, as shown in Figure~\ref{fig:av_loc}.

\vspace{+1mm} \noindent \textbf{\textit{Audio-Visual} Comparative.}
The module focuses on ``instrument,'' ``right,'' ``left,'' and ``louder'' to identify spatial locations in visual modality. Meanwhile, it emphasizes ``louder'' to analyze sound intensities in audio modality. This complementary approach enables the model to tackle the question effectively, as illustrated in Figure~\ref{fig:av_comp}.

\vspace{+1mm} \noindent \textbf{\textit{Audio-Visual} Temporal.}
The attention module balances visual and auditory cues to identify the specific clarinet that produces the first sound. In the visual modality, attention focuses on ``clarinet'' and ``first,'' using motion cues to detect active clarinets. To compensate for visually occluded clarinets, the audio modality emphasizes ``which,'' ``clarinet,'' and ``first,'' helping to highlight the source of the initial sound in Figure~\ref{fig:av_temp}.

\subsubsection{Failure Cases}
\vspace{+1mm} \noindent \textbf{\textit{Audio} Counting.}
The attention incorrectly highlights ``saxophone'' and related auditory features instead of ``trumpet,'' as shown in Figure~\ref{fig:failure_a_count}. Given that only trumpet sounds are present, this misclassification likely stems from the model confusing the trumpet sound with the similar auditory characteristics of a saxophone. Visually, the model also fails to correctly identify ``trumpet,'' possibly due to one being partially obscured and the other blending into the background because of similar coloring with the performers’ clothing. This suggests that the model overly relies on auditory cues when visual distinctions are less prominent, leading to confusion between visually and aurally similar objects.

\vspace{+1mm} \noindent \textbf{\textit{Visual} Location.}
The failure comes from limitations in both modalities. Visually, the absence of the ``bagpipe'' forces reliance on auditory cues. However, with no ``bagpipe'' sound present, overlapping flute and bassoon sounds may have been misclassified, as shown in Figure~\ref{fig:failure_v_loc}. This highlights the challenge of distinguishing similar-sounding instruments in multi-modal reasoning. The issue likely stems from the model’s difficulty in separating distinct auditory features when instrument sounds overlap, compounded by the lack of visual confirmation.

\vspace{+1mm} \noindent \textbf{\textit{Audio-Visual} Temporal.}
Auditorily, the module captures sound-related cues effectively. Visually, attention is drawn to less critical words like ``is,'' which provide some contextual relevance. However, this focus reduces the emphasis on essential keywords such as ``where'' and ``first,'' which are crucial for temporal understanding, as shown in Figure~\ref{fig:failure_av_temp}. Such attention patterns suggest that the model may have overemphasized certain contextual cues while underutilizing spatial and temporal keywords, leading to an incorrect prediction. This indicates a need for better balancing of context and question-specific focus.

\subsection{Qualitative Comparison of Temporal Gaussian}
This section focuses on two main points:
(i) For all nine question types, we demonstrate that the proposed temporal Gaussian approach outperforms conventional sampling methods, such as uniform sampling and Top-K frame selection, by efficiently utilizing the entire temporal sequence and focusing on critical segments in Figure~\ref{fig:good_gaussian_1},~\ref{fig:good_gaussian_2} and ~\ref{fig:good_gaussian_3}. 
(ii) In Figure~\ref{fig:bad_gaussian}, we examine cases where the proposed method underperforms compared to other sampling techniques, offering insights into areas for future improvement.
For comparison, we use ST-AVQA~\cite{music-avqa22} for the uniform sampling and TSPM~\cite{tspm24} for the Top-K frame selection.

\subsubsection{Valid cases}
\vspace{+1mm} \noindent \textbf{\textit{Audio} Counting.}
For the question ``How many musical instruments were heard throughout the video?'' in Figure~\ref{fig:good_gaussian_a}, the Top-K method focuses on selecting frames where musical instruments are most visually prominent. However,  key is to identify segments where audio signals from all playing instruments are strongest. QA-TIGER's audio Gaussian effectively captures these moments, enabling it to count the number of instruments played accurately.

\vspace{+1mm} \noindent \textbf{\textit{Audio} Comparative.}
For a question like ``Is the piano playing longer than the violin?'' in Figure~\ref{fig:good_gaussian_b}, the Top-K approach focuses only on frames that include the piano and violin. However, since it considers only a limited number of $N$ frames, it struggles to make accurate comparisons when the question requires analyzing the entire temporal span, such as for ``longer.'' 
Uniform sampling also fails to consider the entire sequence, making it challenging to derive accurate answers. In contrast, QA-TIGER applies adaptive weights across the entire sequence, allowing for more efficient and accurate comparisons.

\vspace{+1mm} \noindent \textbf{\textit{Visual} Counting.}
QA-TIGER predicts the number of cellos in a video by analyzing both close-up and full-shot scenes of cello performances, as illustrated in Figure~\ref{fig:good_gaussian_c}. In comparison, the uniform sampling may occasionally include frames where all cellos are visible, resulting in a correct prediction. However, if the frame order changes, the prediction could easily be wrong. Meanwhile, the Top-K method focuses only on the close-up frames where cellos are most visible. As a result, it predicts only ``three'' cellos, even though ``four'' are actually being played.

\vspace{+1mm} \noindent \textbf{\textit{Visual} Location.}
The Top-K approach primarily focuses on the ``violin'' itself, selecting frames that are strongly related to the violin but overlooking its right-hand side (Figure~\ref{fig:good_gaussian_d}). While it does select one frame with clues about the instrument to the right of the violin, a close-up of the piano shifts the focus away from this information. In contrast, QA-TIGER effectively concentrates on temporal segments that emphasize both the violin and its right-hand side. It assigns the lowest weight to the piano, ensuring that the most important details are prioritized.

\vspace{+1mm} \noindent \textbf{\textit{Audio-Visual} Existential.}
In Figure~\ref{fig:good_gaussian_e} ``Is the flute in the video always playing?'' highlights a limitation of the Top-K method, which selects only frames where the flute is both present and actively being played. QA-TIGER takes a more balanced approach by allowing the audio Gaussian to focus on moments when the flute is playing, while the visual Gaussian also considers frames where the flute is not being played. This demonstrates that our method enables each modality to independently emphasize different aspects of the question in a complementary manner.

\vspace{+1mm} \noindent \textbf{\textit{Audio-Visual} Counting.}
QA-TIGER focuses on the segments where instruments are being played while broadly considering the overall context, as shown in Figure~\ref{fig:good_gaussian_f}. On the other hand, as observed in other question types, Top-K approach often misses other critical information since it selects only a limited number of frames centered around the specific instrument mentioned in the question. Uniform sampling exclusively uses frames containing only the two individuals playing instruments, leading to the limitation of predicting ``two'' instead of the correct answer, ``four.''

\vspace{+1mm} \noindent \textbf{\textit{Audio-Visual} Location.}
In Figure~\ref{fig:good_gaussian_g}, QA-TIGER's audio Gaussian assigns relatively higher weights to the early parts of the sequence, effectively identifying the first instrument played. Meanwhile, the visual Gaussian focuses on wide shots where the positions of all instruments are visible, contributing to accurately answering the instruments and their locations. However, the uniform sampling approach fails to capture which instrument was played first, and the Top-K method only considers frames with the most prominent close-up of an instrument, limiting its ability to determine the precise location of the instruments.

\vspace{+1mm} \noindent \textbf{\textit{Audio-Visual} Comparative.}
When comparing audio-visual content as in Figure~\ref{fig:good_gaussian_h}, the original video is too short, so the audio for the remaining frames was generated by repeating the last 1-second segment of the video. QA-TIGER accurately focuses only on the relevant segments of the original video across all modalities, enabling precise answer predictions. In contrast, the Top-K method incorrectly focuses on unrelated frames, resulting in wrong answers when paired with uniform sampling.

\vspace{+1mm} \noindent \textbf{\textit{Audio-Visual} Temporal.}
For the question about which congas are played first, as in Figure~\ref{fig:good_gaussian_i}, the Top-K approach focuses on frames highlighting the congas in the early part of the sequence. Unfortunately, due to the limited number of frames, it fails to utilize information about other congas in the later part, leading to an incorrect answer. In contrast, our method effectively reflects the intent of the question by assigning higher weights to the early part through the audio Gaussian, while the visual Gaussian captures critical information such as the total number and locations of congas in the later part, resulting in the correct answer.

\subsubsection{Failure cases}
\vspace{+1mm} \noindent \textbf{\textit{Visual} Counting.}
While QA-TIGER focuses on frames with multiple instruments, it could miss finer details, as shown in Figure~\ref{fig:bad_gaussian_a}.
In the case of the given sample, where ``ukulele'' and ``violin'' coexist, QA-TIGER could struggle to accurately identify each instrument’s appearance or distinguish between them if the audio of one instrument is overshadowed or sounds similar to the other.

\vspace{+1mm} \noindent \textbf{\textit{Audio-Visual} Existential.}
In questions like Figure~\ref{fig:bad_gaussian_b}, ``Is the trumpet in the video always playing?'', QA-TIGER's audio Gaussian effectively focuses on the portions of the audio signal where the trumpet sound is most prominent. However, in the final 5 seconds, the loud cheering from the audience overwhelms the trumpet sound, causing the model to misidentify it and give an incorrect prediction.

\vspace{+1mm} \noindent \textbf{\textit{Audio-Visual} Counting.}
In cases like Figure~\ref{fig:bad_gaussian_c}, the visual Gaussian effectively focuses on the early temporal segment where the ukulele is played, while the audio Gaussian prioritizes the acoustic guitar, which has a stronger audio signal but a similar sound to the ukulele. Although the weight is lower, QA-TIGER's consideration of the entire sequence allows it to include the ukulele sound from the early segment. This enables the model to correctly answer ``two'' for the question, ``How many sounding ukuleles are in the video?''

Overall, these cases suggest opportunities to enhance QA-TIGER, such as implementing adaptive mechanisms to dynamically adjust the number of experts based on content, improving its ability to capture varying temporal complexities. Additionally, addressing external noise can further refine its performance across diverse scenarios.

\section{Discussion \& Future Work}
We provide supplementary material to complement the main paper by detailing experimental setups, additional results, and visualizations of QA-TIGER’s mechanisms. 
QA-TIGER achieves state-of-the-art performance, leveraging Gaussian experts for precise temporal integration and effective alignment of question-specific audio-visual features with minimal redundancy. Compared to prior methods, it demonstrates superior accuracy in complex reasoning tasks, including temporal and comparative queries, while maintaining computational efficiency.
In addition to quantitative improvements, visualizations show how QA-TIGER dynamically adjusts its attention across audio and visual modalities, effectively handling diverse question types.

While QA-TIGER shows promising results both quantitatively and qualitatively, we aim to further enhance its adaptability and flexibility. 
The current Mixture of Experts (MoE) framework relies on a fixed number of experts, which may not fully capture the varying temporal complexities present in different audio-visual content. In this regard, it will be promising to develop adaptive mechanisms that dynamically adjust the number of experts based on the multimodal content, enabling the model to better represent and model varying temporal dynamics. 
Additionally, since AVQA models are often constrained to predefined answers, the integration of large language models into QA-TIGER can be investigated in the future. This integration allows for more flexible and natural language responses, broadening its applicability to more complex and diverse scenarios.

\input{figure/suppl_good_gaussian}

\input{figure/suppl_bad_gaussian}

%% file: algorithm/gaussian_experts.tex
\begin{algorithm}
\caption{Gaussian Experts Module}\label{algo:gaussian_experts}
\begin{algorithmic}
    \State \textbf{Input:} Sentence-level question features: $q_s \in \mathbb{R}^{D}$. Video/Audio features: $\mathbf{v}_q,\mathbf{a}_q \in \mathbb{R}^{T \times D}$. Visual/Audio-related patch features : $\mathbf{p}_v,\mathbf{p}_a \in \mathbb{R}^{T \times D}$. 
    \State \textbf{Output:} Temporal integrated features:  Aggregated visual-related patch features $\tilde{v}_{p_v} \in \mathbb{R}^{D}$, aggregated audio-related patch features $\tilde{v}_{p_a} \in \mathbb{R}^{D}$. Aggregated audio features $\tilde{a} \in \mathbb{R}^{D}$.
    
    \vspace{0.5em}
    \State \textbf{Initialization:} Initialize the center of $E$ experts to the central positions of $E$ segments.
        \State \quad $\text{margin} \gets \frac{1}{2 \times E}$
\Comment{Margin between Gaussian centers}     
        \State \quad $\mathbf{u}_{\text{fixed}} \gets \left[\text{margin} + i \cdot \frac{1 - 2 \cdot \text{margin}}{E - 1} \text{ for } i = 0 \text{ to } E - 1 \right]$
    
    \vspace{0.5em}
    \State \textbf{1. Question-Guided Attention:}
    \State \quad $\mathbf{v}^{\prime}_q, \: \mathbf{a}^{\prime}_q \gets \text{CA}(q_s, \mathbf{v}_q, \mathbf{v}_q),  \: \text{CA}(q_s, \mathbf{a}_q, \mathbf{a}_q)$
    
    \vspace{0.5em}
    \State \textbf{2. Calculate Experts Probability:}
    \State \quad $\mathbf{r}_v, \: \mathbf{r}_a \gets \text{Softmax}(\text{Router}(\mathbf{v}^{\prime}_q)), \: \text{Softmax}(\text{Router}(\mathbf{a}^{\prime}_q))$

    \vspace{0.5em}
    \State \textbf{3. Gaussian Weight Generation:}
    \State \quad $u_{\text{offset}|v}, \sigma_{v} \gets \text{Gaussian Generator}(\mathbf{v}^{\prime}_q)$
    \State \quad $u_{\text{offset}|a}, \sigma_{a} \gets \text{Gaussian Generator}(\mathbf{a}^{\prime}_q)$
    \State \quad Adjust centers and normalize widths:
    \State \quad $u^{i}_{v} \gets u^{i}_{\text{fixed}} + \text{Tanh}(u^{i}_{\text{offset}|v}) \cdot \text{margin}$
    \State \quad $u^{i}_{a} \gets u^{i}_{\text{fixed}} + \text{Tanh}(u^{i}_{\text{offset}|a}) \cdot \text{margin}$
    \State \quad $\sigma^{i}_{v}, \: \sigma^{i}_{a} \gets \text{Sigmoid}(\sigma^{i}_{v}), \: \text{Sigmoid}(\sigma^{i}_{a})$
    \State \quad Generate temporal Gaussian weights:
    \State \quad \textbf{for} $i \gets 1$ to $E$ \textbf{do}
    \State \quad \quad $\mathbf{g}_{v}[i], \: \mathbf{g}_{a}[i] \gets \mathcal{N}(\mathbf{u}^{i}_{v}, (\sigma^{i}_{v})^2), \: \mathcal{N}(\mathbf{u}^{i}_{a}, (\sigma^{i}_{a})^2)$
    \State \quad \quad $\mathbf{g}_{v}[i], \: \mathbf{g}_{a}[i] \gets \frac{\mathbf{g}_{v}[i]}{\max(\mathbf{g}_{v}[i])}, \: \frac{\mathbf{g}_{a}[i]}{\max(\mathbf{g}_{a}[i])}$
    \State \quad \textbf{end for}

    \vspace{0.5em}
    \State \textbf{4. Integration of Experts Output:}
    \State \quad \quad $\mathbf{p}_v = \mathbf{v}_q + \mathbf{p}_v$
    \State \quad \quad $\mathbf{p}_a = \mathbf{v}_q + \mathbf{p}_a$
    \vspace{0.2em}
    \State \quad \quad $\tilde{a} \gets \sum_{i=1}^{E} \mathbf{g}^{i}_{a} \mathbf{r}^{i}_{a} \mathcal{E}^{i}(\mathbf{a}_q)$
    \State \quad \quad $\tilde{v}_{p_v}, \tilde{v}_{p_a} \gets \sum_{i=1}^{E} \mathbf{g}^{i}_{v} \mathbf{r}^{i}_{v} \mathcal{E}^{i}(\mathbf{p}_v), \sum_{i=1}^{E} \mathbf{g}^{i}_{v} \mathbf{r}^{i}_{v} \mathcal{E}^{i}(\mathbf{p}_a)$
    \vspace{0.2em}
    \State \quad \quad \Return $\tilde{v}_{p_v}, \tilde{v}_{p_a}, \tilde{a}$
    
\end{algorithmic}
\end{algorithm}

%% file: table/encoder_comp.tex
\begin{table}[t]
\begin{center}
\begin{adjustbox}{width=\linewidth,center=\linewidth}
\begin{tabular}{l|c|ccc|c}
\hline
\textbf{Method} & 
\textbf{\begin{tabular}[c]{@{}c@{}}CLIP\\Encoder\end{tabular}} & 
\textbf{\begin{tabular}[c]{@{}c@{}}A-QA\end{tabular}} & 
\textbf{\begin{tabular}[c]{@{}c@{}}V-QA\end{tabular}} & 
\textbf{\begin{tabular}[c]{@{}c@{}}AV-QA\end{tabular}} & 
\textbf{\begin{tabular}[c]{@{}c@{}}Avg\end{tabular}} \\ 
\hline
PSTP-Net~\cite{pstp23} & B/32 & 70.91 & 77.26 & 72.57 & 73.52  \\
TSPM~\cite{tspm24}     & B/32 & \textbf{76.91} & 81.92 & 72.57 & 75.81  \\ 
\textbf{QA-TIGER}      & B/32 & 76.66 & \textbf{83.69} & \textbf{72.61} & \textbf{76.26} \\ 
\hline
\hline
PSTP-Net~\cite{pstp23} & L/14 & 73.87 & 79.19 & 71.76 & 74.10  \\
TSPM~\cite{tspm24}     & L/14 & 76.91 & 83.61 & 73.51 & 76.79  \\ 
\textbf{QA-TIGER}      & L/14 & \textbf{78.58} & \textbf{85.14} & \textbf{73.74} & \textbf{77.62}  \\ 
\hline
\end{tabular}
\end{adjustbox}
\caption{Results for different Encoders (CLIP-B/32 and CLIP-L/14) used for both visual and textual feature extraction.}
\label{tab:encoder_comp}
\vspace{-7mm}
\end{center}
\end{table}

%% file: figure/visual_gaussian_example.tex
\begin{figure}[t]
    \centering
    \begin{subfigure}[t]{1.0\linewidth}
        \centering
        \includegraphics[width=1.0\textwidth]{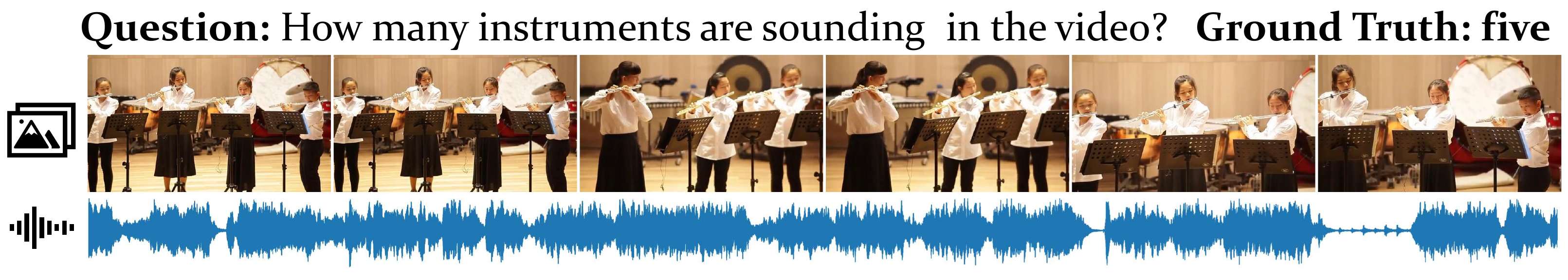}
        \subcaption{Input sample}
        \label{gaussian_input}
    \end{subfigure}
    \begin{subfigure}[t]{1.0\linewidth}
        \centering
        \includegraphics[width=1.0\textwidth]{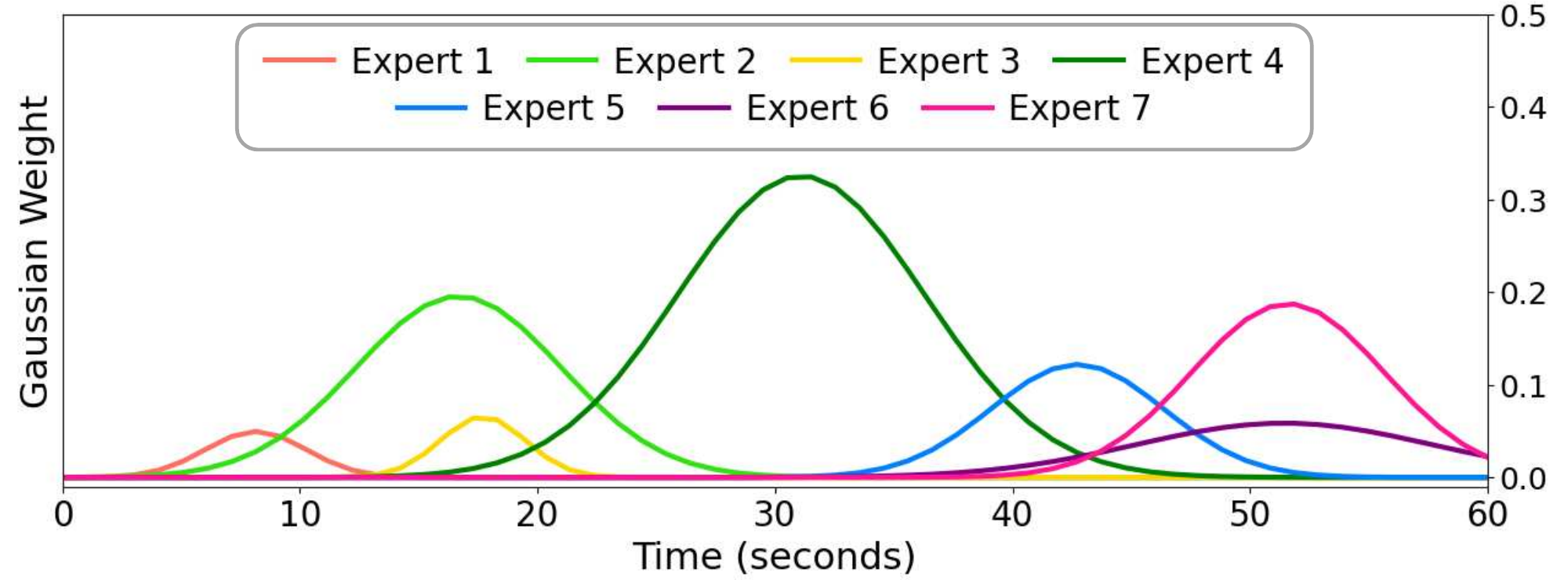}
        \subcaption{Temporal visual weights from Gaussian experts}
        \label{visual_experts}
    \end{subfigure}

    \begin{subfigure}[t]{1.0\linewidth}
        \centering
        \includegraphics[width=1.0\textwidth]{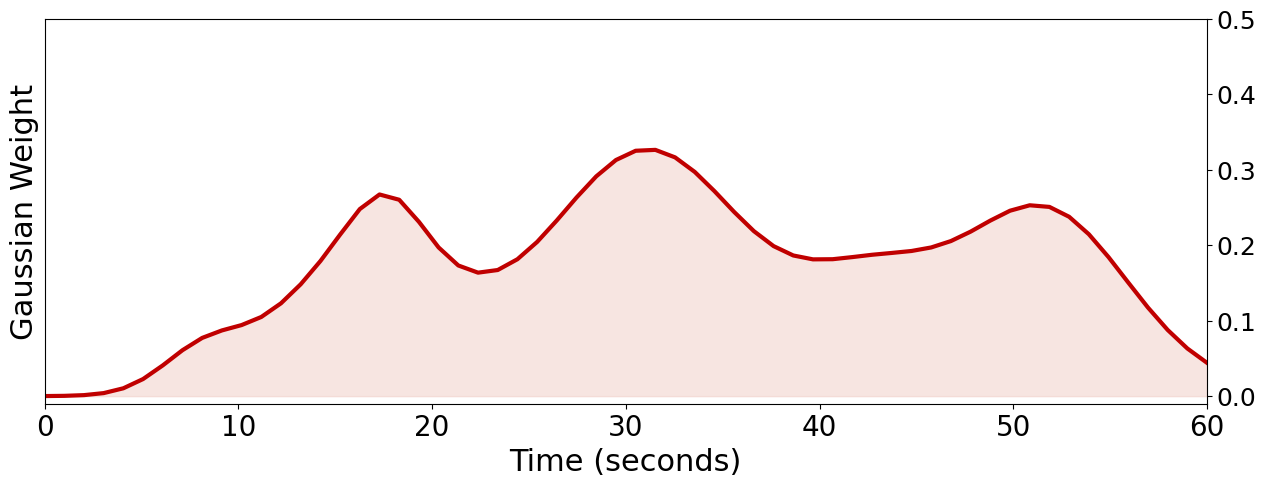}
        \subcaption{Integrated visual Gaussian weights}
        \label{visual_gaussian}
    \end{subfigure}

    \begin{subfigure}[t]{1.0\linewidth}
        \centering
        \includegraphics[width=1.0\textwidth]{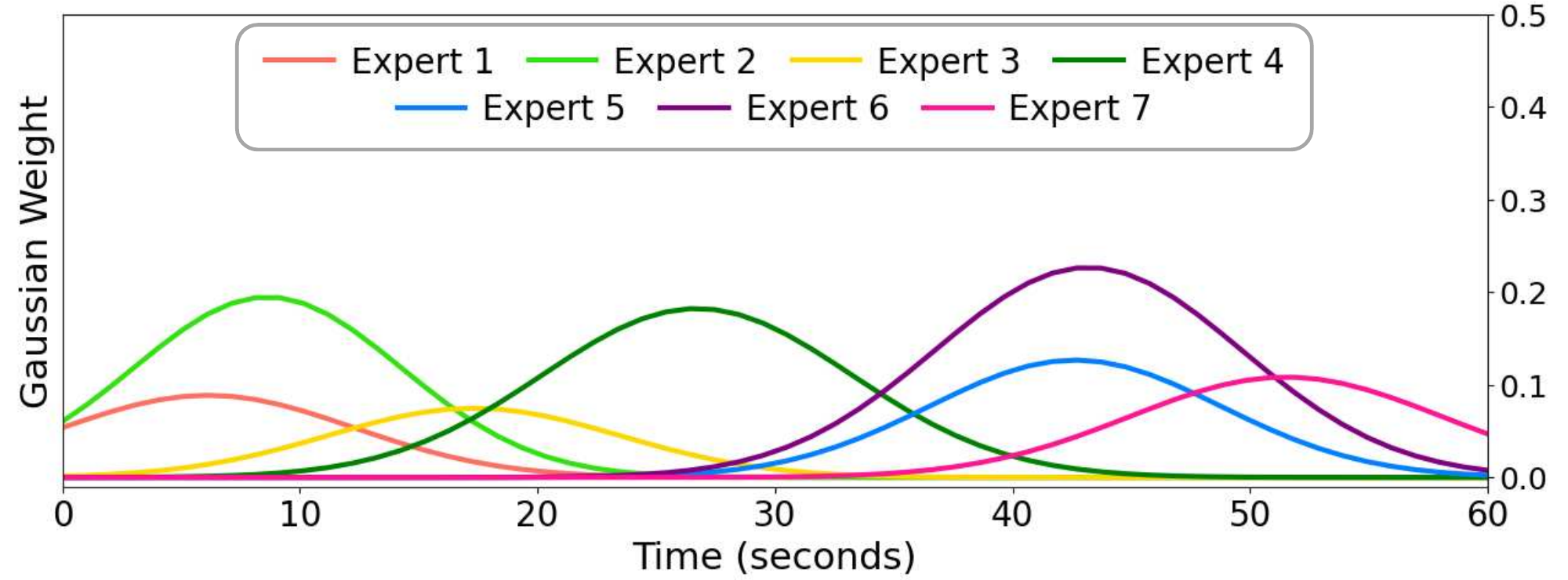}
        \subcaption{ Temporal audio weights from Gaussian experts}
        \label{audio_experts}
    \end{subfigure}

    \begin{subfigure}[t]{1.0\linewidth}
        \centering
        \includegraphics[width=1.0\textwidth]{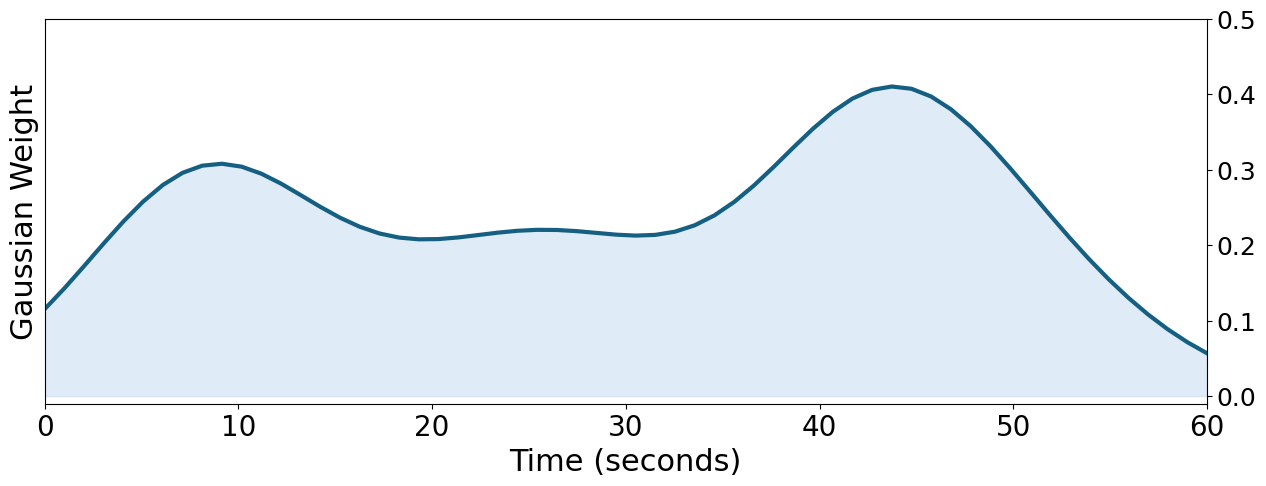}
        \subcaption{Integrated audio Gaussian weights}
        \label{audio_gaussian}
    \end{subfigure}
    \caption{Visualization of temporal weights from Gaussian experts for visual and audio modalities, integrated to focus on question-relevant frames for accurate predictions.}
    \label{fig:gaussian_example}
    \vspace{-3mm}
\end{figure}

%% file: figure/supple_pos_a_attention.tex
\begin{figure*}[t]
    \centering
    \begin{subfigure}[t]{0.85\linewidth}
        \centering
        \includegraphics[width=\linewidth]{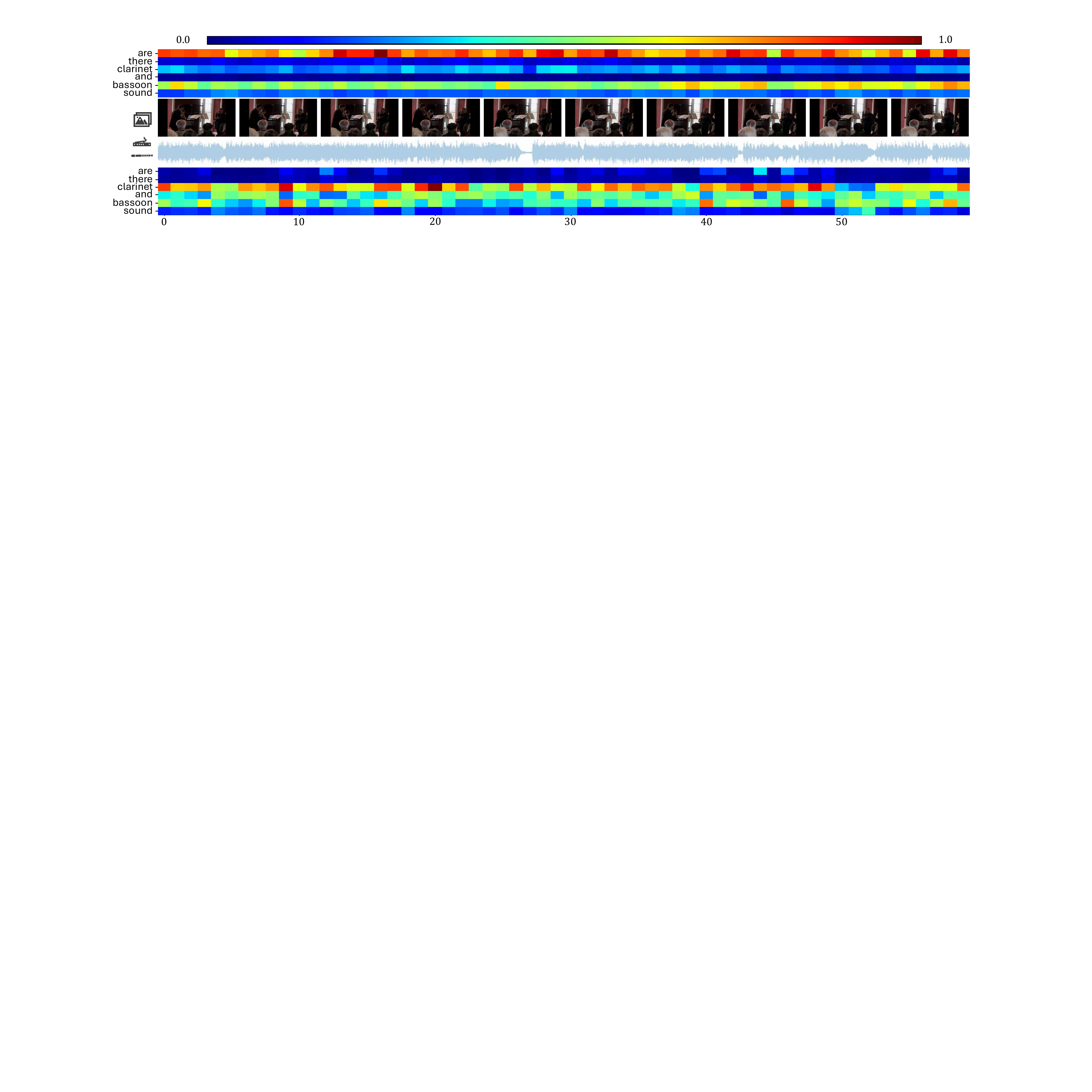}
        \vspace{-5mm}
        \subcaption{Audio Counting}
        \label{fig:audio_counting}
    \end{subfigure}
    \vspace{3mm} %
    \begin{subfigure}[t]{0.85\linewidth}
        \centering
        \includegraphics[width=\linewidth]{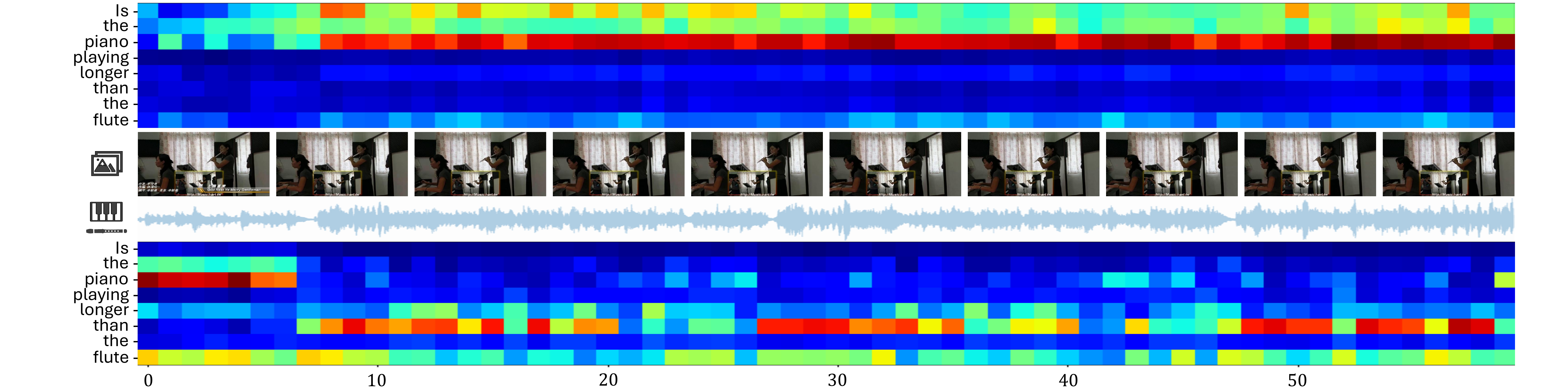}
        \vspace{-5mm}
        \subcaption{Audio Comparative}
        \label{fig:audio_comparison}
    \end{subfigure}
    \vspace{-7mm}
    \caption{Valid attention visualization for audio questions.}
    \label{fig:positive_attention}
    \vspace{-3mm}
\end{figure*}

%% file: figure/supple_pos_v_attention.tex
\begin{figure*}[t]
    \centering
    \begin{subfigure}[t]{0.85\linewidth}
        \centering
        \includegraphics[width=\linewidth]{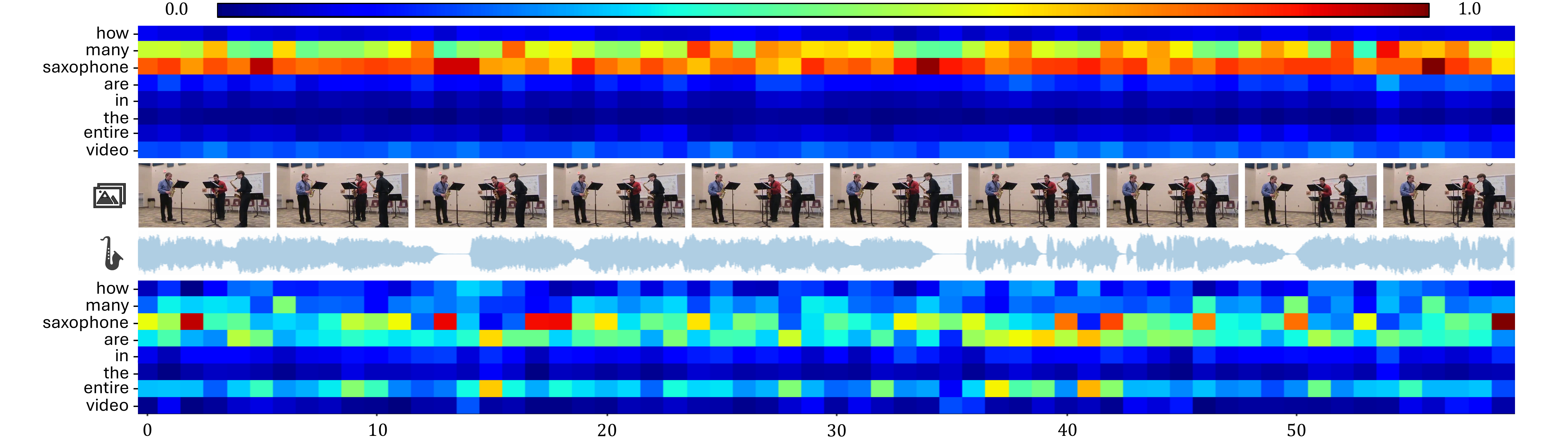}
        \vspace{-5mm}
        \subcaption{Visual Counting}
        \label{fig:visual_counting}
    \end{subfigure}
    \vspace{3mm} %
    \begin{subfigure}[t]{0.85\linewidth}
        \centering
        \includegraphics[width=\linewidth]{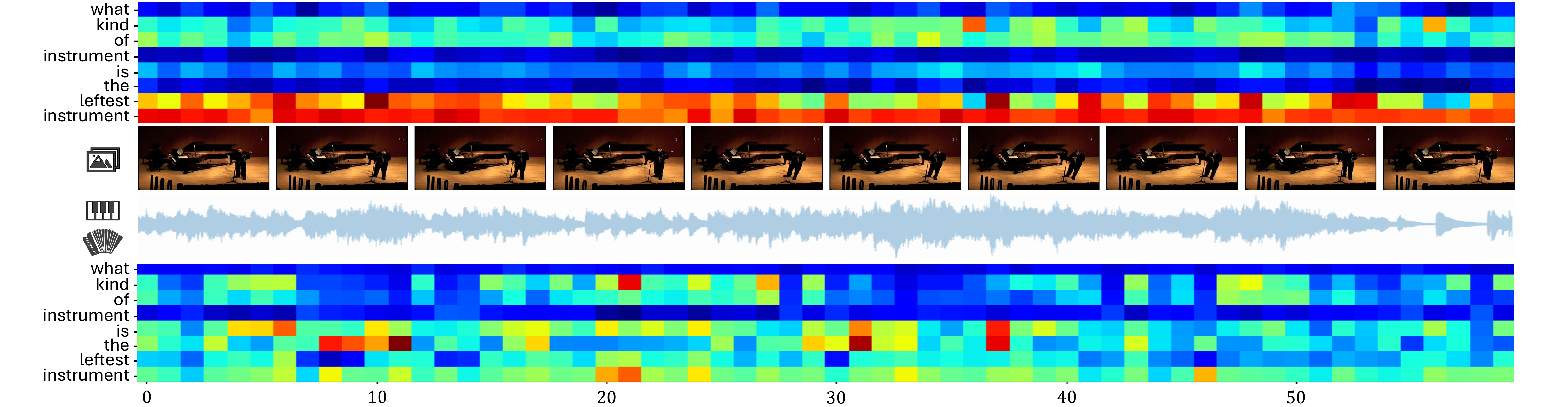}
        \vspace{-5mm}
        \subcaption{Visual Location}
        \label{fig:visual_location}
    \end{subfigure}
    \vspace{-7mm}
    \caption{Valid attention visualization for visual questions.}
    \label{fig:positive_attention_visual}
    \vspace{-3mm}
\end{figure*}

%% file: figure/supple_pos_av_attention.tex
\begin{figure*}[!htbp]
    \centering
    \begin{subfigure}[t]{0.85\linewidth}
        \centering
        \includegraphics[width=\linewidth]{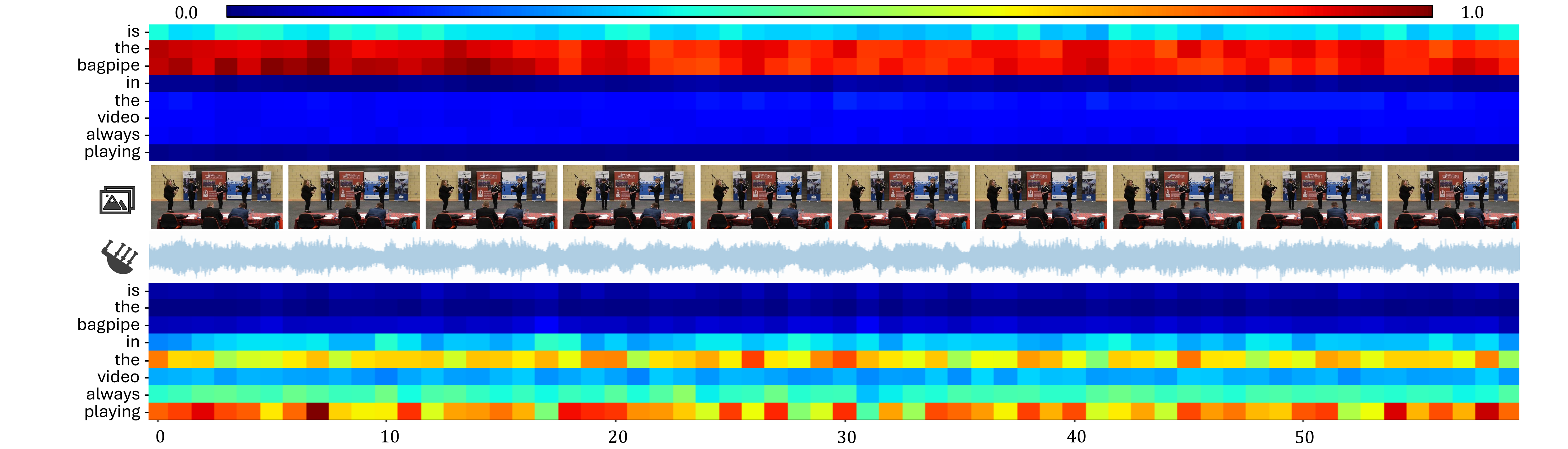}
        \vspace{-5.5mm}
        \subcaption{Audio-Visual Existence}
        \label{fig:av_exist}
    \end{subfigure}
    \begin{subfigure}[t]{0.85\linewidth}
        \centering
        \includegraphics[width=\linewidth]{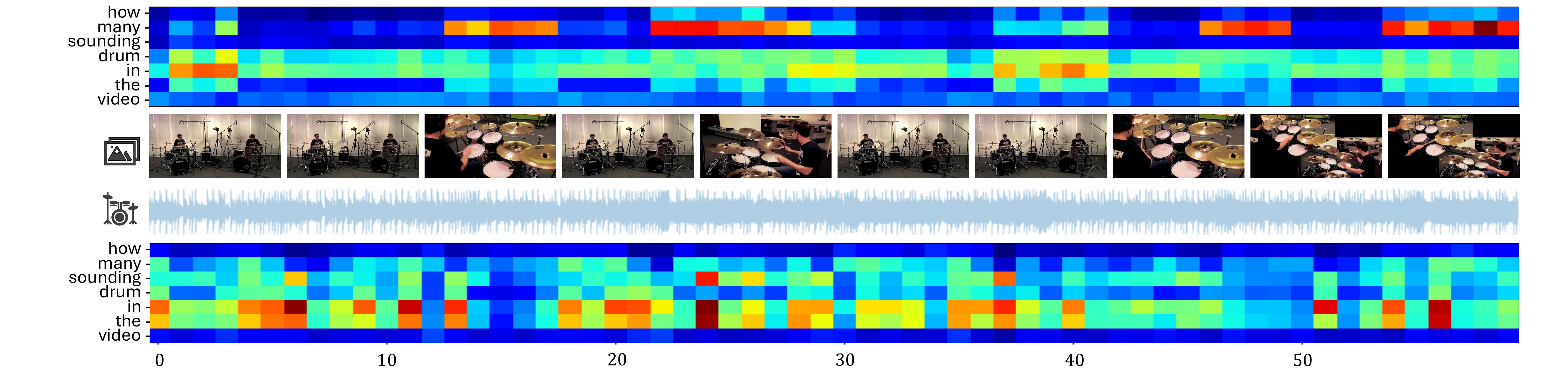}
        \vspace{-5.5mm}
        \subcaption{Audio-Visual Counting}
        \label{fig:av_count}
    \end{subfigure}
    \begin{subfigure}[t]{0.85\linewidth}
        \centering
        \includegraphics[width=\linewidth]{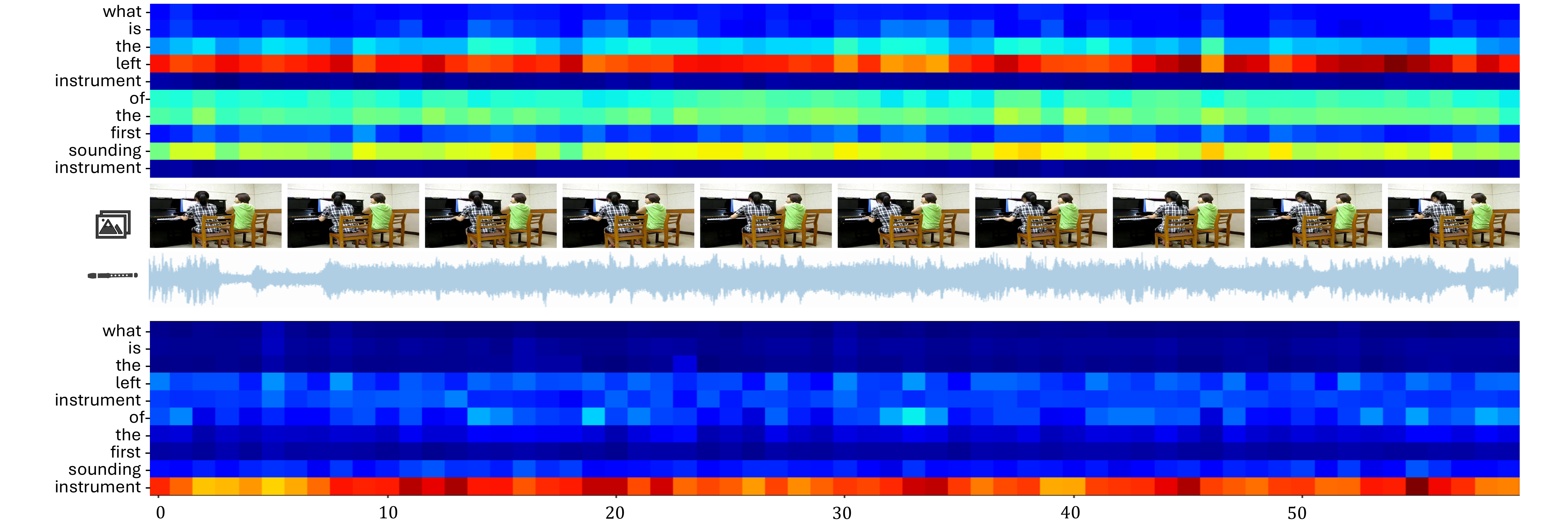}
        \vspace{-5.5mm}
        \subcaption{Audio-Visual Location}
        \label{fig:av_loc}
    \end{subfigure}
    \begin{subfigure}[t]{0.85\linewidth}
        \centering
        \includegraphics[width=\linewidth]{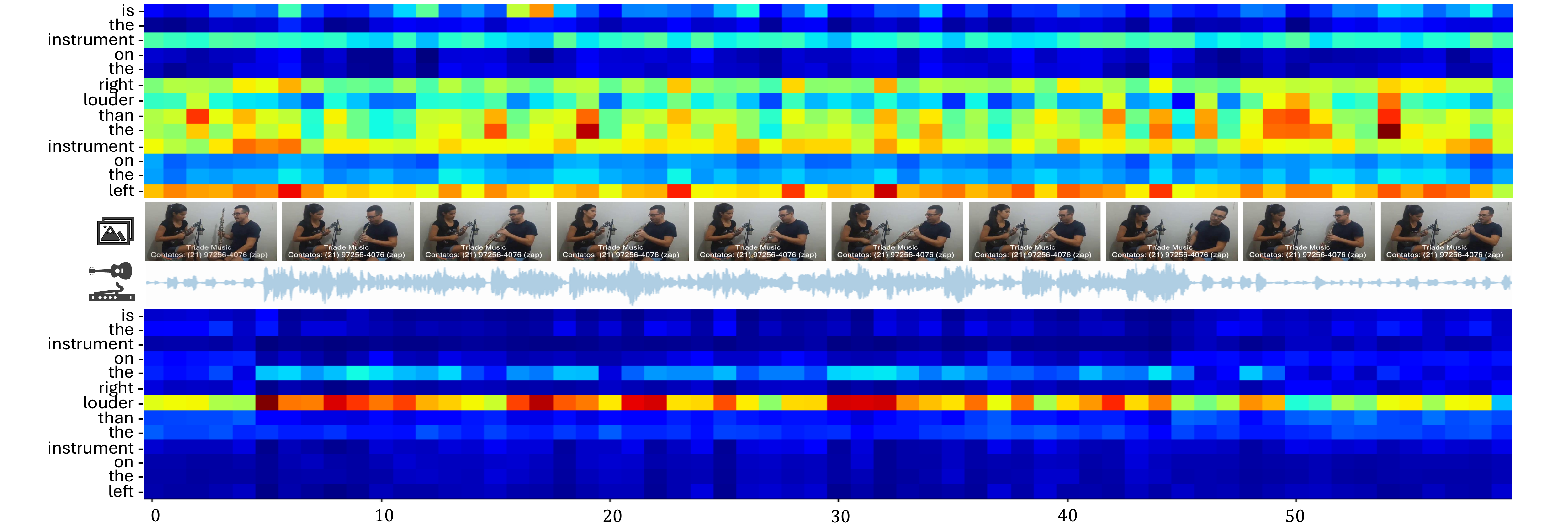}
        \vspace{-5.5mm}
        \subcaption{Audio-Visual Comparative}
        \label{fig:av_comp}
    \end{subfigure}
    \begin{subfigure}[t]{0.85\linewidth}
        \centering
        \includegraphics[width=\linewidth]{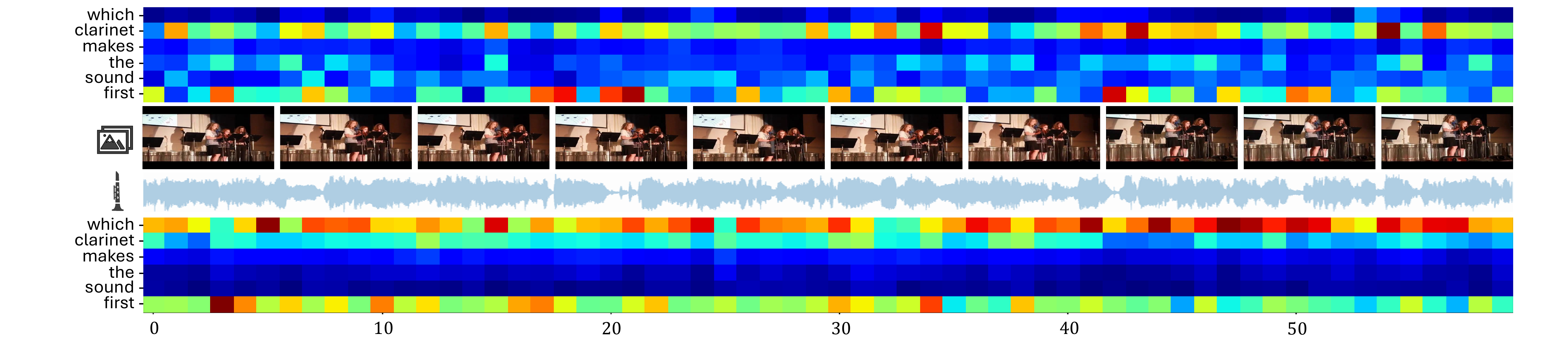}
        \vspace{-5.5mm}
        \subcaption{Audio-Visual Temporal}
        \label{fig:av_temp}
    \end{subfigure}
    \vspace{-3mm}
    \caption{Valid attention visualization for audio-visual type.}
    \label{fig:positive_attention_av}
    \vspace{-3mm}
\end{figure*}

%% file: figure/suppl_neg_attention.tex
\begin{figure*}[ht]
    \centering
    \begin{subfigure}[t]{0.85\linewidth}
        \centering
        \includegraphics[width=\linewidth]{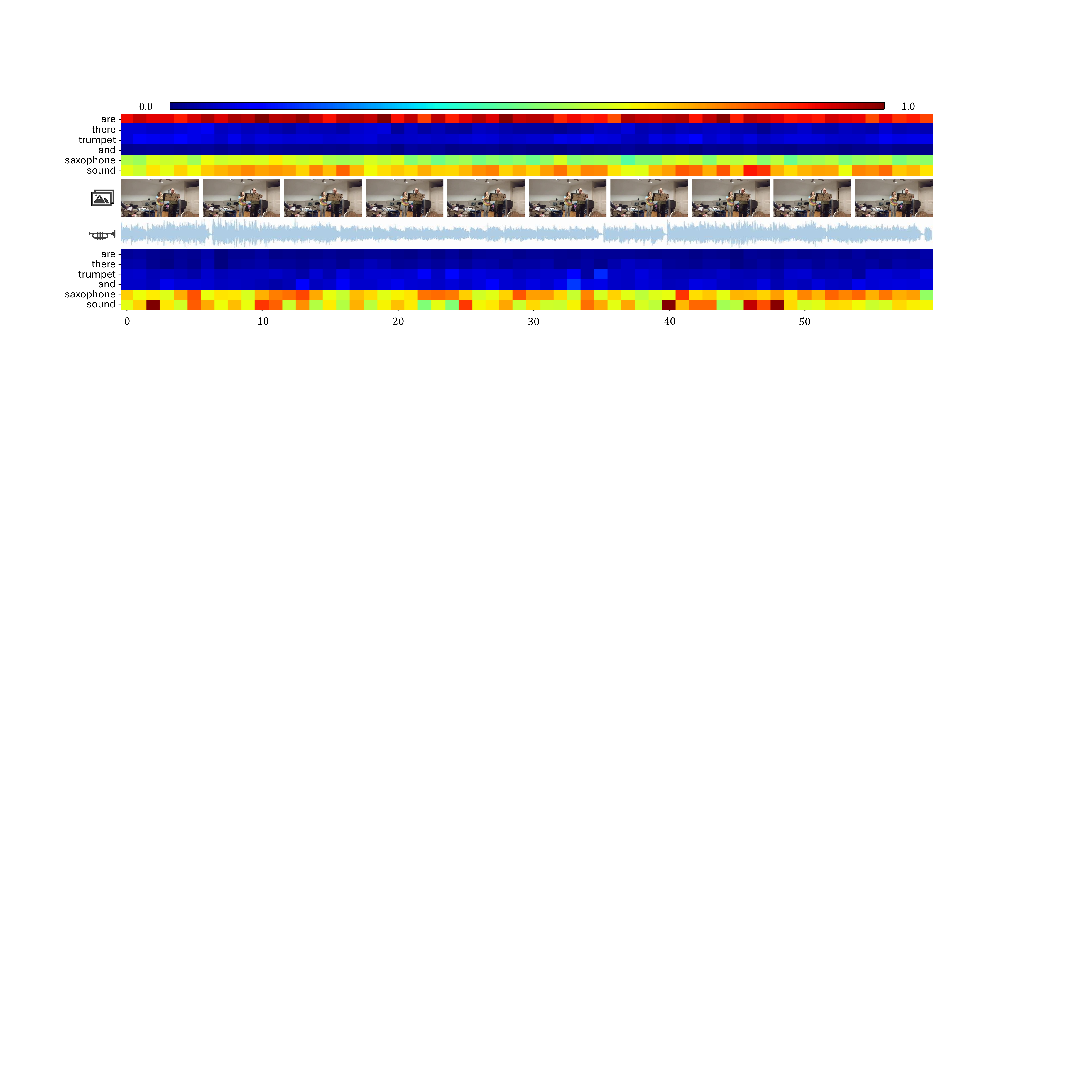}
        \subcaption{Audio Counting}
        \label{fig:failure_a_count}
    \end{subfigure}
    \vspace{2mm} %
    \begin{subfigure}[t]{0.85\linewidth}
        \centering
        \includegraphics[width=\linewidth]{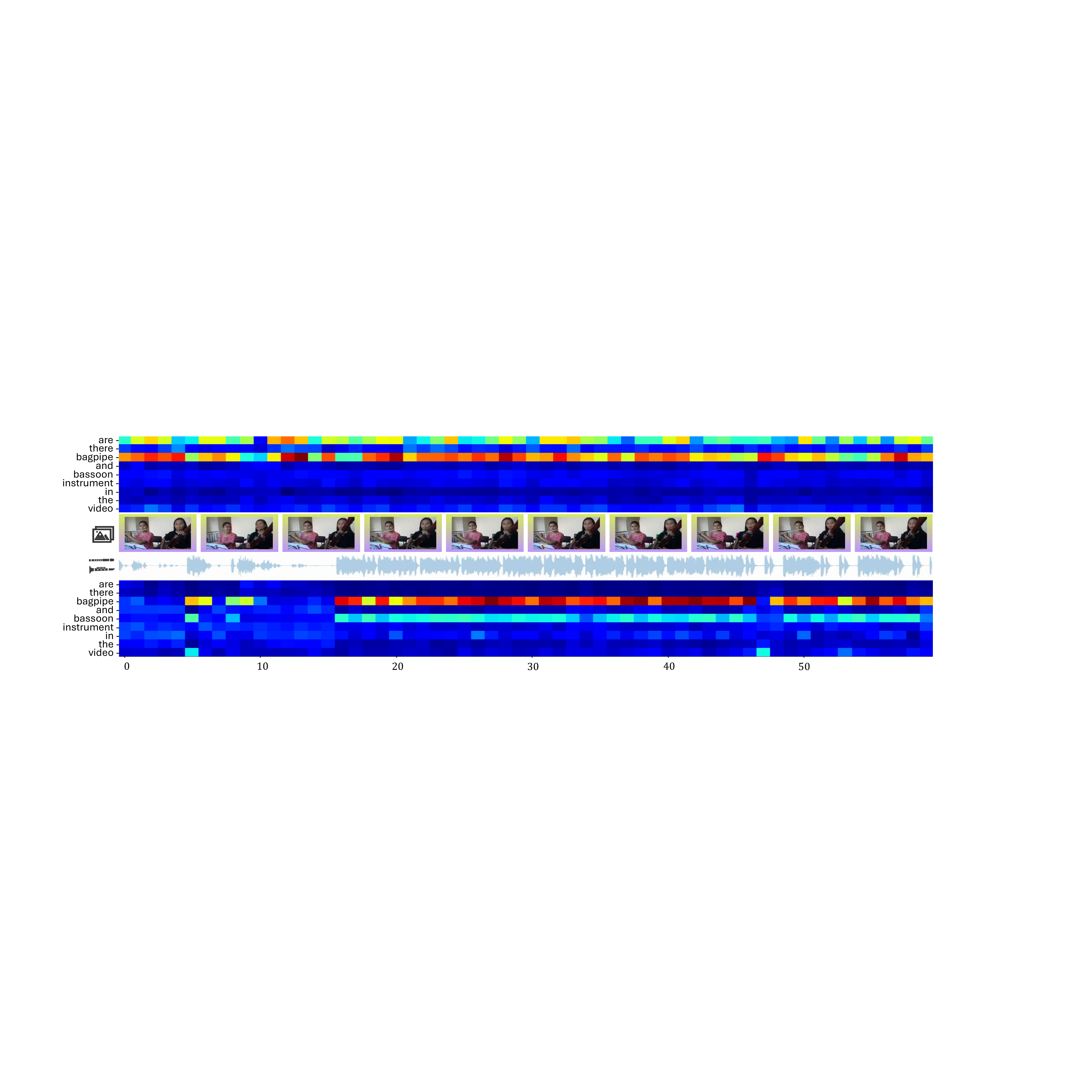}
        \subcaption{Visual Location}
        \label{fig:failure_v_loc}
    \end{subfigure}
    \vspace{2mm}
    \begin{subfigure}[t]{0.85\linewidth}
        \centering
        \includegraphics[width=\linewidth]{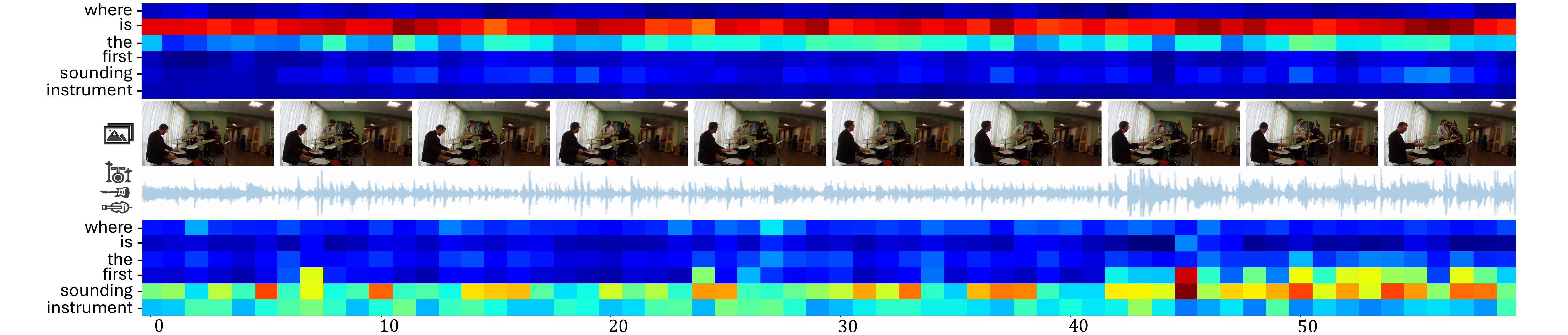}
        \subcaption{Audio-Visual Temporal}
        \label{fig:failure_av_temp}
    \end{subfigure}
    \vspace{-5mm}
    \caption{Attention visualization in failure cases.}
    \label{fig:negative_attention}
\end{figure*}

%% file: figure/suppl_good_gaussian.tex
\begin{figure*}[t]
    \vspace{2cm}
    \centering
    \setlength{\tabcolsep}{30pt} %
    \renewcommand{\arraystretch}{5} %
    \begin{tabular}{ccc} %
        \rotatebox{90}{ %
            \begin{subfigure}[b]{1.2\linewidth}
                \centering
                \includegraphics[width=1.1\linewidth]{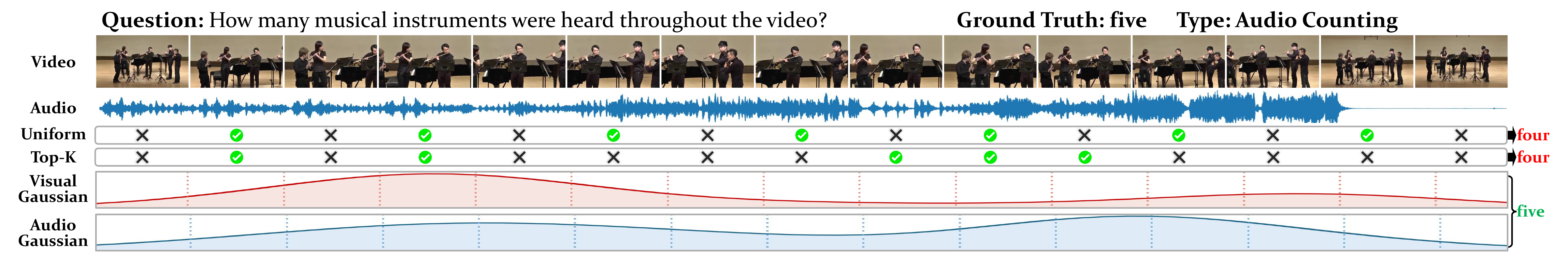}
                \caption{Audio Counting}
                \label{fig:good_gaussian_a}
            \end{subfigure}
        } &
        \rotatebox{90}{
            \begin{subfigure}[b]{1.2\linewidth}
                \centering
                \includegraphics[width=1.1\linewidth]{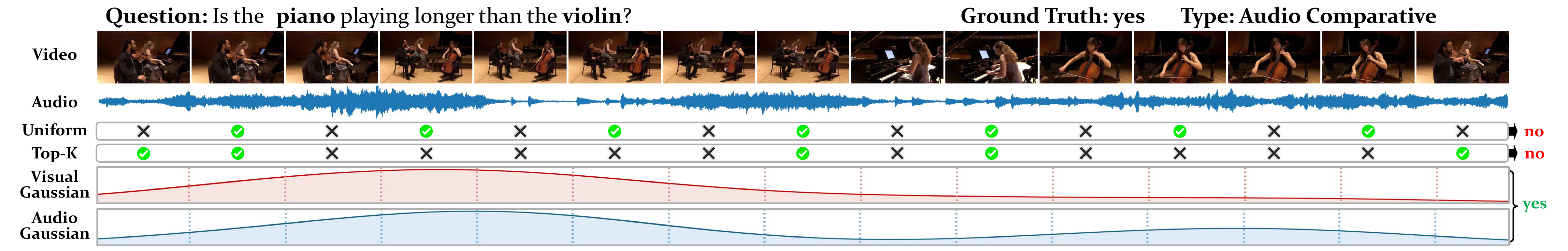}
                \caption{Audio Comparative}
                \label{fig:good_gaussian_b}
            \end{subfigure}
        } &
        \rotatebox{90}{
            \begin{subfigure}[b]{1.2\linewidth}
                \centering
                \includegraphics[width=1.1\linewidth]{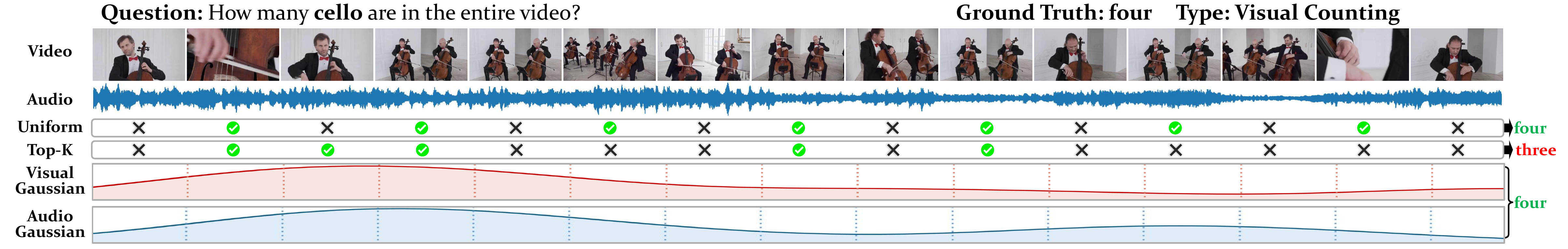}
                \caption{Visual Counting}
                \label{fig:good_gaussian_c}
            \end{subfigure}
        }
    \end{tabular}
    \vspace{-3mm}
    \caption{Valid qualitative comparison with Uniform sampling and Top-K frame selection.}
    \label{fig:good_gaussian_1}
\end{figure*}

\clearpage

\begin{figure*}[t]
    \vspace{2cm}
    \centering
    \setlength{\tabcolsep}{30pt} %
    \renewcommand{\arraystretch}{5} %
    \begin{tabular}{ccc} %
        \rotatebox{90}{ %
            \begin{subfigure}[b]{1.2\linewidth}
                \centering
                \includegraphics[width=1.1\linewidth]{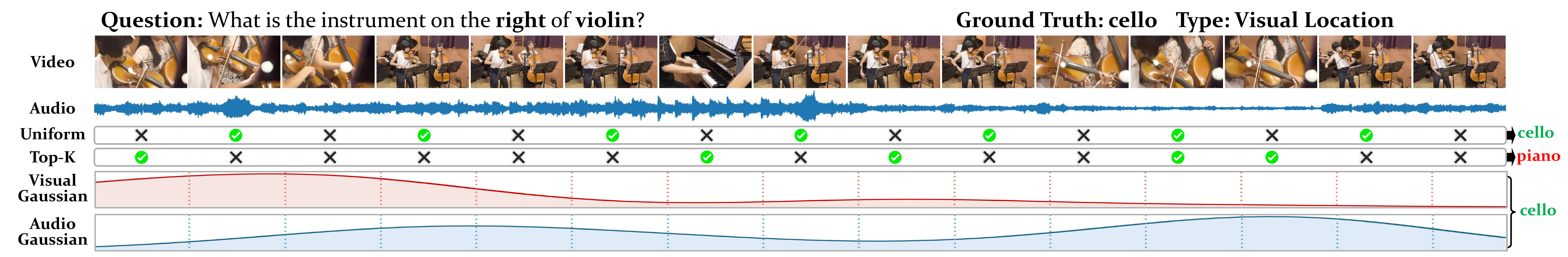}
                \caption{Visual Location}
                \label{fig:good_gaussian_d}
            \end{subfigure}
        } &
        \rotatebox{90}{
            \begin{subfigure}[b]{1.2\linewidth}
                \centering
                \includegraphics[width=1.1\linewidth]{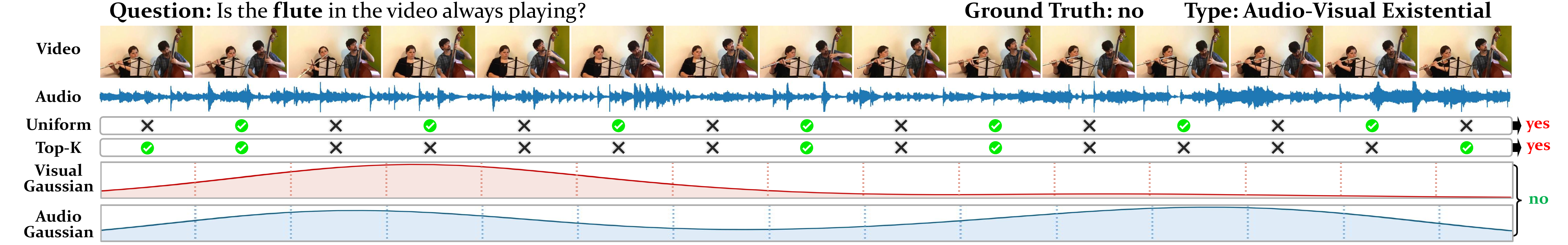}
                \caption{Audio-Visual Existential}
                \label{fig:good_gaussian_e}
            \end{subfigure}
        } &
        \rotatebox{90}{
            \begin{subfigure}[b]{1.2\linewidth}
                \centering
                \includegraphics[width=1.1\linewidth]{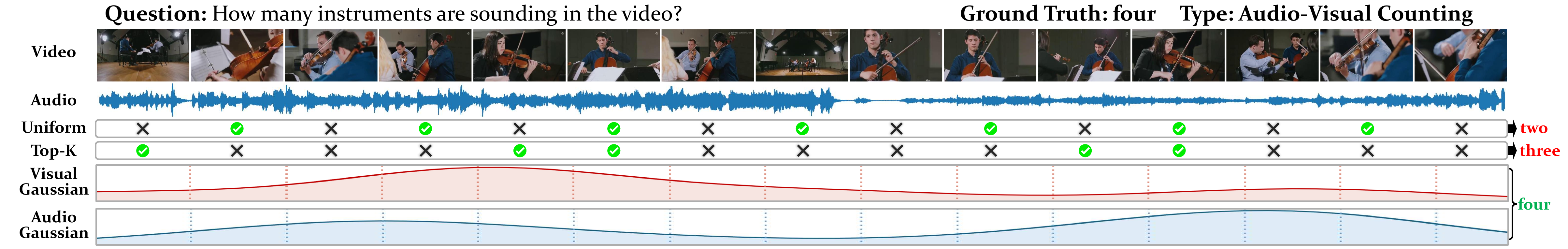}
                \caption{Audio-Visual Counting}
                \label{fig:good_gaussian_f}
            \end{subfigure}
        }
    \end{tabular}
    \vspace{-3mm}
    \caption{Valid qualitative comparison with Uniform sampling and Top-K frame selection.}
    \label{fig:good_gaussian_2}
\end{figure*}

\clearpage

\begin{figure*}[t]
    \vspace{2cm}
    \centering
    \setlength{\tabcolsep}{30pt} %
    \renewcommand{\arraystretch}{5} %
    \begin{tabular}{ccc} %
        \rotatebox{90}{ %
            \begin{subfigure}[b]{1.2\linewidth}
                \centering
                \includegraphics[width=1.1\linewidth]{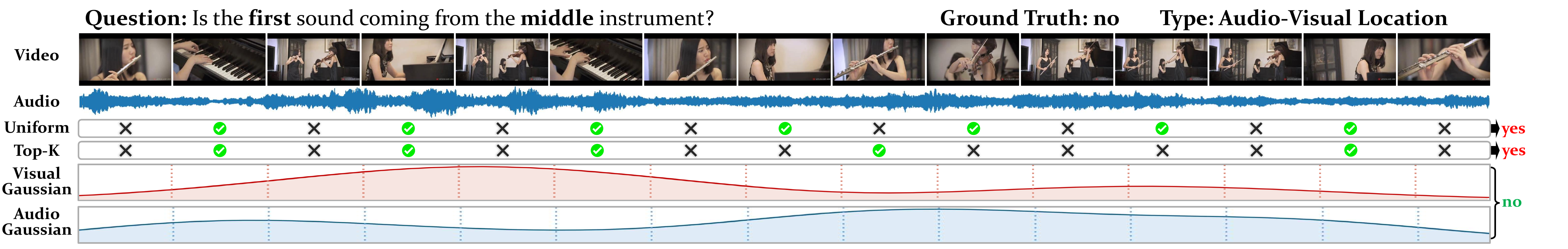}
                \caption{Audio-Visual Location}
                \label{fig:good_gaussian_g}
            \end{subfigure}
        } &
        \rotatebox{90}{
            \begin{subfigure}[b]{1.2\linewidth}
                \centering
                \includegraphics[width=1.1\linewidth]{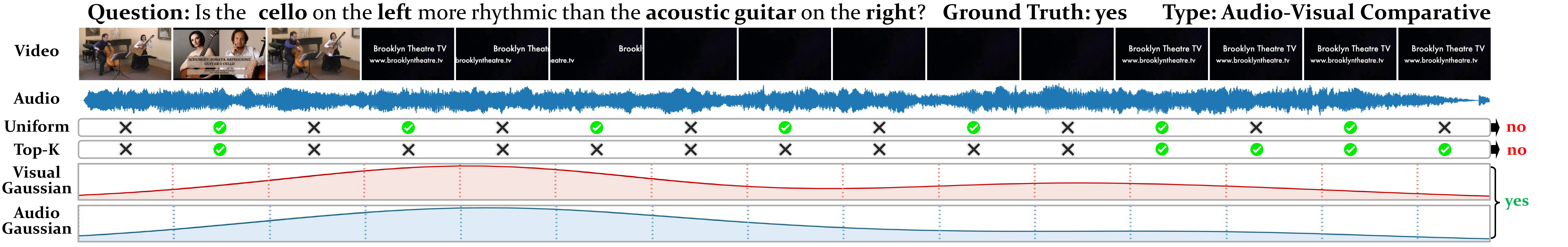}
                \caption{Audio-Visual Comparative}
                \label{fig:good_gaussian_h}
            \end{subfigure}
        } &
        \rotatebox{90}{
            \begin{subfigure}[b]{1.2\linewidth}
                \centering
                \includegraphics[width=1.1\linewidth]{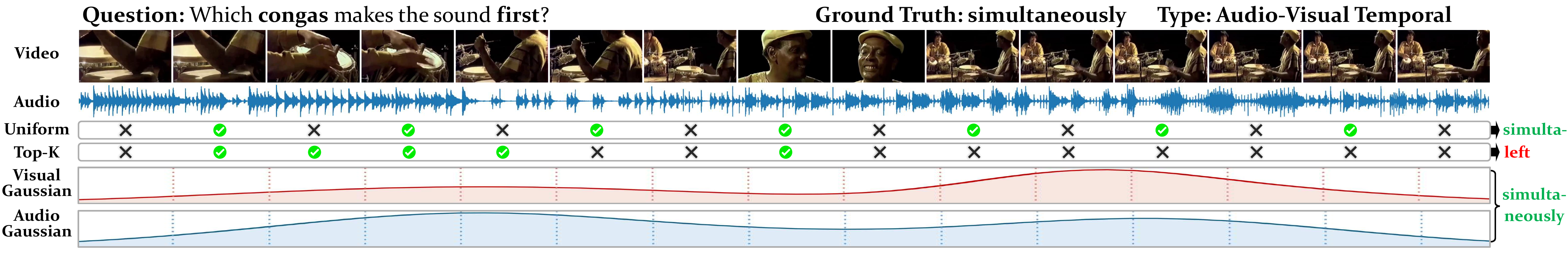}
                \caption{Audio-Visual Temporal}
                \label{fig:good_gaussian_i}
            \end{subfigure}
        }
    \end{tabular}
    \vspace{-3mm}
    \caption{Valid qualitative comparison with Uniform sampling and Top-K frame selection.}
    \label{fig:good_gaussian_3}
\end{figure*}

%% file: figure/suppl_bad_gaussian.tex
\clearpage

\begin{figure*}[t]
    \vspace{2cm}
    \centering
    \setlength{\tabcolsep}{30pt} %
    \renewcommand{\arraystretch}{5} %
    \begin{tabular}{ccc} %
        \rotatebox{90}{ %
            \begin{subfigure}[b]{1.2\linewidth}
                \centering
                \includegraphics[width=1.1\linewidth]{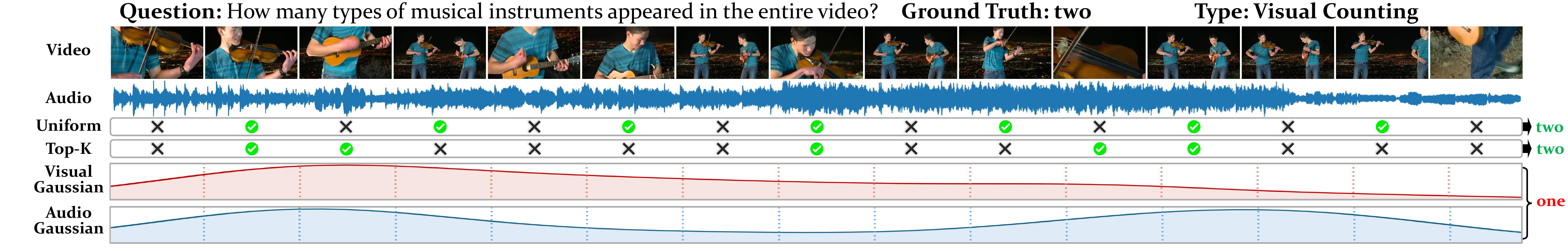}
                \caption{Visual Counting}
                \label{fig:bad_gaussian_a}
            \end{subfigure}
        } &
        \rotatebox{90}{
            \begin{subfigure}[b]{1.2\linewidth}
                \centering
                \includegraphics[width=1.1\linewidth]{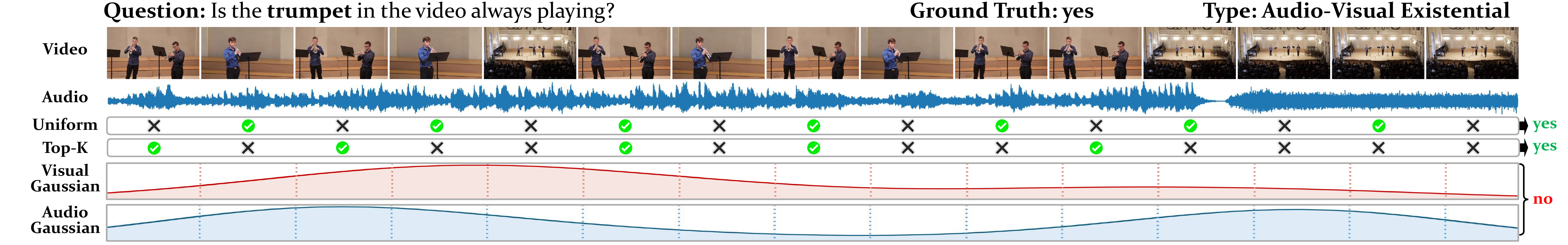}
                \caption{Audio-Visual Existential}
                \label{fig:bad_gaussian_b}
            \end{subfigure}
        } &
        \rotatebox{90}{
            \begin{subfigure}[b]{1.2\linewidth}
                \centering
                \includegraphics[width=1.1\linewidth]{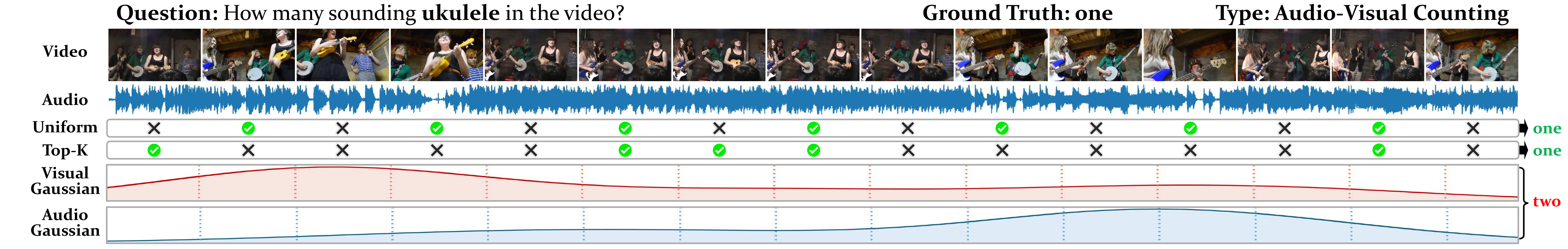}
                \caption{Audio-Visual Counting}
                \label{fig:bad_gaussian_c}
            \end{subfigure}
        }
    \end{tabular}
    \vspace{-3mm}
    \caption{Failure qualitative comparison with Uniform sampling and Top-K frame selection.}
    \label{fig:bad_gaussian}
\end{figure*}